\documentclass[10pt,twocolumn,letterpaper]{article}

\usepackage{cvpr}              
\usepackage[accsupp]{axessibility} 
\usepackage{graphicx}
\usepackage{amsmath}
\usepackage{amssymb}
\usepackage{booktabs}
\usepackage{latexsym}
\usepackage{algorithm}
\usepackage{algorithmic}
\usepackage{colortbl}
\usepackage{dsfont}
\usepackage{stackengine}
\usepackage{xcolor}
\usepackage{hyperref}
\usepackage{multicol}
\usepackage{multirow}

\usepackage[capitalize]{cleveref}
\crefname{section}{Sec.}{Secs.}
\Crefname{section}{Section}{Sections}
\Crefname{table}{Table}{Tables}
\crefname{table}{Tab.}{Tabs.}

\def\cvprPaperID{4574} 
\def\confName{CVPR}
\def\confYear{2022}

\begin{document}

\title{Towards Robust Adaptive Object Detection under Noisy Annotations}

\author{Xinyu Liu$^{1}$ \qquad Wuyang Li$^{1}$ \qquad Qiushi Yang$^{1}$ \qquad Baopu Li$^{2}$ \qquad Yixuan Yuan$^{1,}$\thanks{Corresponding author. This work was supported by Hong Kong Research Grants Council (RGC) General Research Fund 11211221 (CityU 9043152).}\\
$^{1}$City University of Hong Kong \qquad $^{2}$Baidu USA LLC\\
{\tt\small \{xliu423-c, wuyangli2-c, qsyang2-c\}@my.cityu.edu.hk, baopuli@baidu.com} \\
{\tt\small yxyuan.ee@cityu.edu.hk}}
\maketitle

\begin{abstract}
   Domain Adaptive Object Detection (DAOD) models a joint distribution of images and labels from an annotated source domain and learns a domain-invariant transformation to estimate the target labels with the given target domain images. Existing methods assume that the source domain labels are completely clean, yet large-scale datasets often contain error-prone annotations due to instance ambiguity, which may lead to a biased source distribution and severely degrade the performance of the domain adaptive detector de facto. In this paper, we represent the first effort to formulate noisy DAOD and propose a Noise Latent Transferability Exploration (NLTE) framework to address this issue. It is featured with 1) Potential Instance Mining (PIM), which leverages eligible proposals to recapture the miss-annotated instances from the background; 2) Morphable Graph Relation Module (MGRM), which models the adaptation feasibility and transition probability of noisy samples with relation matrices; 3) Entropy-Aware Gradient Reconcilement (EAGR), which incorporates the semantic information into the discrimination process and enforces the gradients provided by noisy and clean samples to be consistent towards learning domain-invariant representations. A thorough evaluation on benchmark DAOD datasets with noisy source annotations validates the effectiveness of NLTE. In particular, NLTE improves the mAP by 8.4\% under 60\% corrupted annotations and even approaches the ideal upper bound of training on a clean source dataset. \footnote{Code is available at \href{https://github.com/CityU-AIM-Group/NLTE}{https://github.com/CityU-AIM-Group/NLTE}.}
\end{abstract}

\section{Introduction}
\label{sec:intro}

\begin{figure}[t]
	\centering
	\footnotesize
   \begin{tabular}{c@{\hskip0.1pt}c@{\hskip0.1pt}c}
   \includegraphics[width=2.62cm]{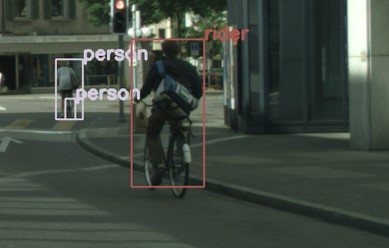}&%
		\includegraphics[width=2.62cm]{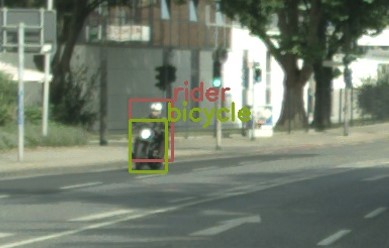}&%
		\includegraphics[width=2.62cm]{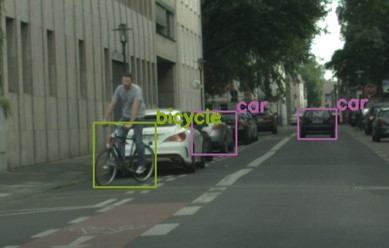}\\
		(a) & (b) & (c) \\
	\end{tabular}\vspace{-8pt}
	  \caption{Examples of noisy annotations in Cityscapes dataset. \textbf{Miss-annotated} samples: The \textit{bicycle} in (a); the \textit{rider} and \textit{car} in (c). \textbf{Class-corrupted} samples: The \textit{rider} and \textit{bicycle} are labeled as \textit{person} in (a); the \textit{motorcycle} is labeled as \textit{bicycle} in (b).}
	\label{fig:teaser}
	\vspace{-12pt}
\end{figure}

Recent years have witnessed great progress in domain adaptive object detection (DAOD) \cite{DAfasterrcnn, swda, atf, zheng2020cross, xu2020crossgraph, sapn, everypixelmatters, zhao2020adaptive, wu2021vector}. It alleviates the performance drop of the detectors when applied to unseen domains due to the domain shift.
Most DAOD methods are constructed with domain adversarial training \cite{grl}, in which a domain classifier is proposed to train the feature extractor to perform a domain-invariant transformation of images from different domains.
However, existing methods are all built with an ideal condition that a clean source domain is accessible, which is impractical in many real-world applications \cite{bg_med, bg_ridar}. 
The annotations can be noisy due to various reasons, including ambiguous objects caused by occlusion or obscureness, limited crowd-sourcing or labeling time, low quality labeled web-crawled images, etc. \cite{web0, web1}
Frustratingly, the noisy class annotations occur frequently, even in benchmark DAOD source datasets such as Cityscapes, as shown in Fig. \ref{fig:teaser}. The noisy annotations can be categorized into two groups: \textit{miss-annotated} instances (Fig. \ref{fig:teaser} (a), (c)) and \textit{class-corrupted} instances (Fig. \ref{fig:teaser} (a), (b)).
More specifically, it has been studied that addressing the classification error is critical to the detector \cite{borji2019empiricalupperbound, zhang2021varifocalnet}, thus the noisy class labels in the source dataset could severely damage the domain adaptive detectors. 


\begin{figure}[t]
  \centering
  \vspace{-8pt}
    \includegraphics[width=0.8\linewidth]{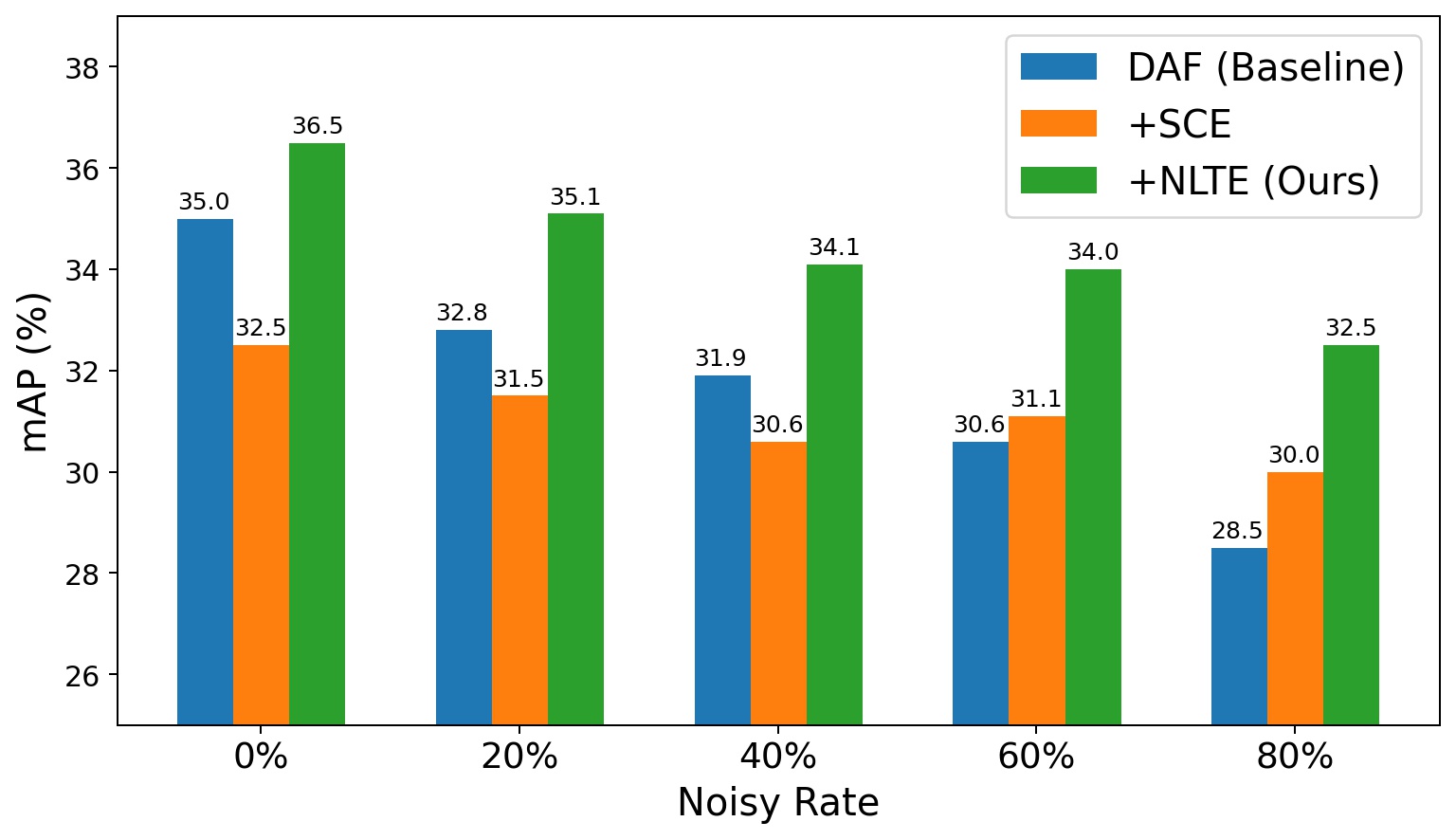}
   \vspace{-9pt}
   \caption{Performance comparison of baseline domain adaptive detector DAF \cite{DAfasterrcnn}, adding noise-robust learning method SCE \cite{sce}, and adding the proposed NLTE under different noisy rates in Pascal VOC \& Noisy Pascal VOC $\to$ Clipart1k.}
   \label{fig:NLTE_SCE_COMPARE}
   \vspace{-14pt}
\end{figure}

The intuitive solution for solving the noisy DAOD problem is to combine approaches in learning with noisy labels for classification and domain adaptive object detection. However, this direct combination may encounter several challenges. 
Firstly, existing methods in learning with noisy labels for image classification \cite{cp, sce, guo2018curriculumnet} minimize or totally filter out the impact of noisy annotated samples during training the network. While in DAOD, the source images with noisy labels are still useful for aligning with target domain as the domain discriminator is class-agnostic, and the target images could benefit source dataset denoising in reverse. If source samples with rich domain-specific information are filtered out by these methods, the adaptation process will be seriously affected \cite{yu2020label}.
Secondly, these methods are designed for datasets that are corrupted by class-conditional noise between foreground categories with approximately balanced distributions \cite{sce, gce, APL, SR}, while noisy DAOD contains diversified noise and imbalanced foreground-background ratio, thus is more complicated and intractable via these methods.
Finally, the essentially differed optimization perspective between detection and classification (\textit{i.e.}, the detection task requires multiple losses to work synergistically) \cite{li2020towards} makes it non-trival to extend existing noise-robust approaches into DAOD frameworks. Hence they could suffer from underfitting \cite{APL} and require elaborately tuning when adopted to the detection framework. We adopt the noise-robust learning approach SCE \cite{sce} into the domain adaptive object detector DAF \cite{DAfasterrcnn} and display the results in Fig. \ref{fig:NLTE_SCE_COMPARE}. Although mAPs are improved under large noisy rates, SCE severely affects the detector under clean scenario and low noisy rates, which is undesirable for real world applications.

%
To address the critical yet undeveloped noisy DAOD issue, we propose a novel Noise Latent Transferability Exploration (NLTE) framework to simultaneously address the negative impact caused by \textit{miss-annotated} and \textit{class-corrupted} samples, which facilitates the training of domain adaptive object detectors under noisy source annotations. 
Specifically, to mine potential \textit{miss-annotated} samples for enriching the source semantic features, we propose Potential Instance Mining (PIM), which looks into the background proposals and dynamically recaptures eligible instances according to their prediction uncertainty.
Secondly, considering that the latent domain-related knowledge and semantic information of \textit{class-corrupted} samples are important and attainable for domain alignment, we propose a Morphable Graph Relation Module (MGRM) to leverage their transferable representations for progressively enhancing the discrimination ability of the model. We first conduct intra-domain feature aggregation with morphable graphs, then generate a global relation matrix which is updated by the aggregated node features to model the class-wise transition probability across domains. Afterwards, local relation matrices are built to explore the alignment feasibility of noisy samples, and the transition probabilities of noisy samples are regularized by the global relation matrix.
Finally, as both \textit{miss-annotated} and \textit{class-corrupted} noises are contributive to learn domain-invariant representations, we propose an Entropy-Aware Gradient Reconcilement (EAGR) strategy for harmonizing the adaptation procedure of noisy and clean samples. It affiliates class confidence into the discriminator, then enforces the gradients of clean and noisy samples to be consistent towards a domain-invariant direction.
Experiments are conducted on both synthetic noisy datasets 
and real-world scenarios. NLTE outperforms various possible baselines, which validates its effectiveness.
With $60\%$ noisy rate, NLTE can significantly improve the mAP of the baseline domain adaptive detector by $8.4\%$, and only drops by $~2\%$ when compared with the clean scenario.

\section{Related Works}
\label{sec:relatedworks}

\subsection{Learning with Noisy Annotations}

To train a robust model under noisy annotations, different methods have been proposed and can be approximately divided into three categories. 
The first category is loss correction or adjustment methods \cite{Coteaching, correction2, correction3, correction4, correction5, correction6, cp}. 
They tried to adjust the loss for each training sample or discard unreliable samples. 
However, they are prone to false corrections which may further affect the training process, and will suffer from limited eligible data under a high noisy rate. 
The second is to design symmetric losses that are robust to noise \cite{robustloss, sce, gce, APL, SR}. 
Generalized Cross Entropy (GCE) \cite{gce} combined the merit of MAE \cite{robustloss} and cross entropy loss. 
Symmetric Cross Entropy (SCE) \cite{sce} utilized a weighted summation of the cross entropy loss and the reverse cross entropy loss to make the classification loss symmetric. 
However, they may only capable to handle certain noisy rates and could collapse under clean scenarios, meanwhile need arduous tuning when applied to the detection task which requires multi-task learning.
The last category is to learn a noise transition matrix to rectify the predictions with extra network components \cite{ntm0, ntm1, ntm2, ntm3}. 
Xia \etal \cite{ntm1} approximated the instance-dependent matrix for an instance by a combination of the matrix for the parts of the instance. 
Li \etal \cite{ntm2} consistently estimated the transition
matrices without anchor points. 
However, these methods are based on the assumption that the labels have strong correlations and are designed for ad-hoc situations, such that they are not suitable for the noisy DAOD scenario. 

\subsection{Domain Adaptive Object Detection}

Unsupervised domain adaptive object detection has been widely utilized for narrowing down the domain gaps between labeled source data and unlabelled target data \cite{DAfasterrcnn, swda, maf, zheng2020cross, atf, htcn, xu2020crossgraph, everypixelmatters, zhao2020adaptive, wu2021vector}. 
Chen \etal \cite{DAfasterrcnn} initially used image-level and instance-level adaptations jointly to reduce the domain gap.
Xu \etal \cite{xu2020crossgraph} aligned local prototypes and designed a class-reweighted contrastive loss for bridging the domain gap.
Zhao \etal \cite{zhao2020adaptive} designed an auxiliary multi-label learning branch for incorporating class to ensure consistent category information between domains.
Wu \etal \cite{wu2021vector} disentangled domain-invariant and domain-specific representations based on vector decomposition. 
However, all previous works assume the source data is clean thus the source category information is reliable for adaptation, while we propose to achieve domain adaptive object detection with a noisy annotated source dataset. 

\subsection{Learning with Noisy Annotations for Instance Recognition}

Chadwick and Newman \cite{chadwick2019training} improved the co-teaching \cite{Coteaching} framework for training the object detector with noisy labels. 
Yang \etal \cite{noisyinstancesegmentation} described different roles of noisy class labels in the instance segmentation task and used cross entropy or symmetric losses to train them. 
Li \etal \cite{li2020towards} utilized the outputs of two diverged classifiers to rectify the bounding boxes and class labels, then trained the model with the corrected labels. 
However, the previous works attempted to train noise-robust detectors within the same domain, while we propose to tackle the problem of performance deterioration caused by the noisy annotations in the source dataset in the DAOD setting and achieve robust domain adaptive detection.
\section{Methodology}
\begin{figure*}[t]
  \centering
    \includegraphics[width=0.90\linewidth]{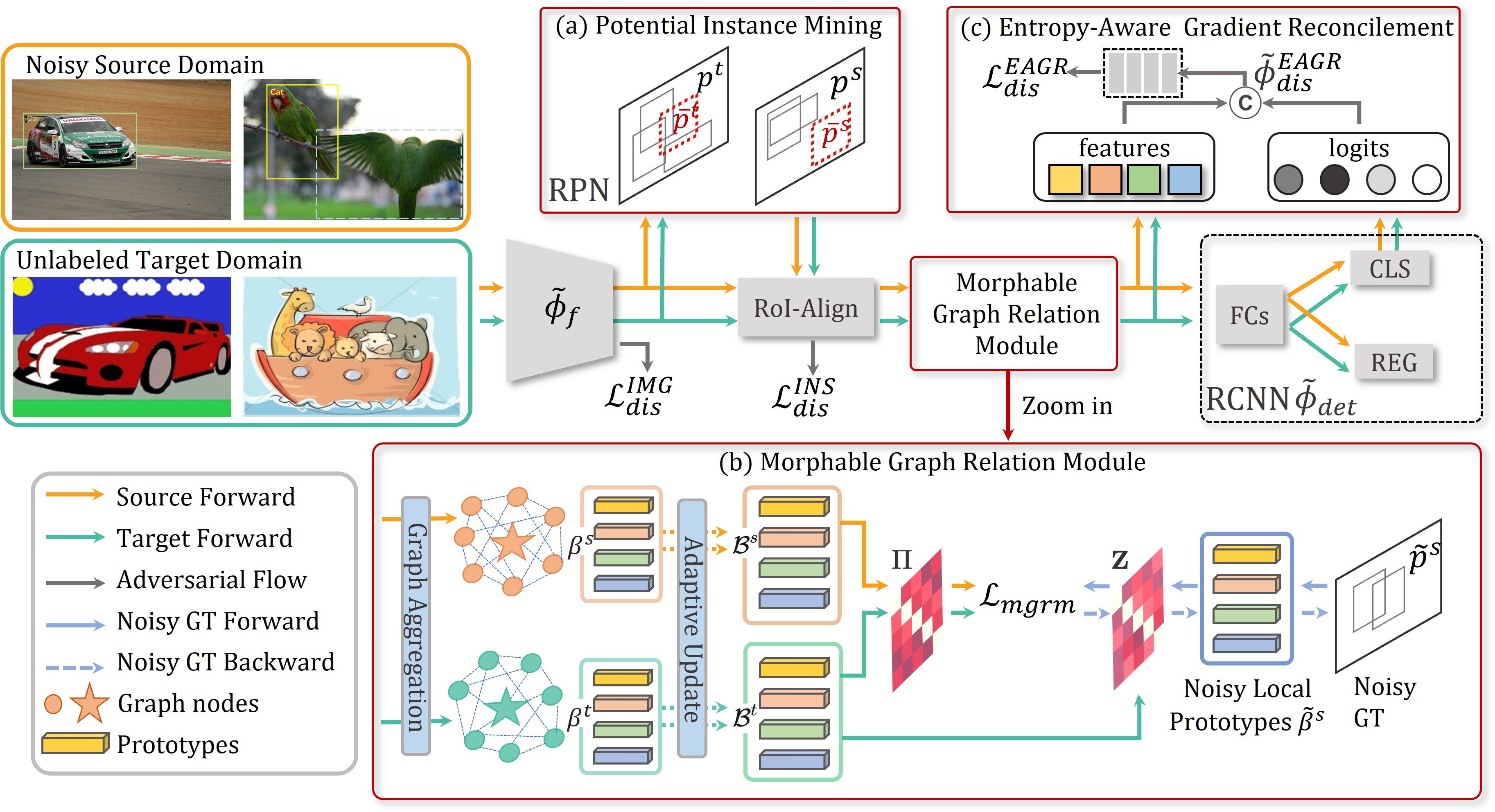}
   \vspace{-10pt}
   \caption{Overview of our NLTE framework, which includes PIM, MGRM and EAGR. \textcircled{c} is the concatenation operation.}
   \label{fig:framework}
   \vspace{-8pt}
\end{figure*}
\subsection{Overview}

\noindent\textbf{Problem setting.} In noisy DAOD, we are given noisy labeled source dataset $\mathcal{D}^s=\{x^s_i, (y^s, \Tilde{y}^s)_i\}_{i=1}^{N_{s}}$ and unlabeled target dataset $\mathcal{D}^t=\{x^t_i\}_{i=1}^{N_t}$. The labels $y^s, \Tilde{y}^s$ both contain box and class annotations, and source and target domains share an identical label space. However, the class annotations in $\Tilde{y}^s$ contain noise.
The objective is to learn a domain-adaptive object detector that can detects objects in $\mathcal{D}^t$.

\noindent\textbf{How label noise affects domain adaptive object detection.} For a domain adaptive detector, it attempts to learn a transformation $\upsilon$ that aligns the conditional distributions of image variable $X$ and label variable $Y$ from different domains, 
such that $P^{{t}}_{\upsilon(X^{{t}}) \mid Y^{{t}}}=P^{{s}}_{\upsilon(X^{{s}}) \mid Y^{{s}}}$. Then with the $P^{{s}}_{X^{{s}}Y^{{s}}}$ estimated from $\mathcal{D}^s$ and the distribution of the drawn samples from $P^{s}_{X^s}$, $P^{t}_{X^t}$, we are able to estimate $P^{t}_{Y^t}$, which is the marginal class distribution in the target domain. However, if the source dataset is noisy, then $P^{t}_{Y^t}$ is computed based on a biased joint distribution $\tilde{P}^{{s}}_{X^{{s}}\Tilde{Y}^{{s}}}$ and the learned transformation $\Tilde{\upsilon}$ will degrade the performance of domain adaptive object detection.

\noindent\textbf{Concept of the proposed method.} Although correcting all noisy labels in the source domain could fully recover the detection performance, it is unattainable practically and may not be optimal for achieving effective domain adaptation. To this end, we build a detection framework that jointly mines miss-annotated samples for recapturing missing semantics and explores the intrinsic positive impact on improving the generalization ability of the detector for class-corrupted samples rather than intuitively correcting the noisy annotations, which is illustrated in Fig. \ref{fig:framework}.

\subsection{Potential Instance Mining}

As miss-annotated samples may cause semantic deficiency and limited domain-invariant representations, we propose PIM to recapture potential foreground instances from background in virtue of the Region Proposal Network (RPN). 
As RPN is class-agnostic, the predicted objectness score of each proposal represents the uncertainty of the existence of an object within the proposal.
Therefore, if the proposals have larger objectness scores than thresholds and no intersection with the ground truth boxes, we select them as eligible candidate proposals $\overline{\mathbf{P}}^s$:
\vspace{-5pt}
\begin{equation}
    \overline{\mathbf{P}}^s = \{\overline{p}_i \ \lvert \ obj(\overline{p}_i)>\tau, \overline{p}_i \notin \mathbf{P}^s, \forall_j IoU(\overline{p}_i, p_j)=0\},
\end{equation}
where $\tau$ is the threshold. PIM is also utilized in the target domain to mine confident positive samples $\overline{\mathbf{P}}^t$ for more effective domain alignment. 
Through the PIM mechanism, only highly-confident proposals are preserved such that missing objects would get recaptured, which simultaneously increases the number of correctly labeled instances for enhancing the discrimination ability and enriches the diversity of source semantic features.

\subsection{Morphable Graph Relation Module}

To explore the embedded domain knowledge and semantic information within class-corrupted samples, we propose MGRM to model the adaptation feasibility and transition probability of these samples. It regularizes the category-wise relations between noisy local prototypes and global prototypes with morphable graphs. The graphs are built upon features from original proposals generated by RPN $\mathbf{P}^s, \mathbf{P}^t$ and proposals explored by PIM $\overline{\mathbf{P}}^s, \overline{\mathbf{P}}^t$. We omit the domain superscript for clearer explanation if the operations are conducted on both domains.

\noindent\textbf{Intra-domain graph feature aggregation.} Given proposals $\mathbf{P} \in \mathbb{R}^{N \times D} \leftarrow \{\mathbf{P}, \overline{\mathbf{P}}\}$ 
after PIM, we first construct them as intra-domain undirected graphs $\mathcal{G} = \{\mathcal{V}, \mathcal{E}\}$.
Specifically, the vertices correspond to the proposals within each domain, and the edges are defined as the feature cosine similarity between them (${e}_{ii'} = \frac{p_i \cdot p_{i'}}{\parallel p_i \parallel_2 \cdot \parallel p_{i'} \parallel_2}$).
Afterwards, we apply intra-domain aggregation to enhance the feature representation within each domain, shown as follows:
\vspace{-6pt}
\begin{equation}
    {p}_i \leftarrow \sigma\Big(\sum_{i' \in {\rm Neighbour}(i)}(w_{i}p_i{e}_{ii'} + p_i)\Big), \quad p_i \in \mathbf{P},
\vspace{-6pt}
\end{equation}
where the Neighbour($i$) denotes the proposals within the same domain as $i$, $\sigma$ is an activation function, and $w_{i}$ is a learnable weight that maps the original feature dimension $D$ to $D'$. After the aggregation, proposals that share the common features could improve the feature representation and generate robust proposal features $\mathbf{P}$ with enhanced adaptive contexts within each domain.

\noindent\textbf{Global relation matrix construction.} 
To model the semantic relationship and effectively explore the category-wise transition probability between source and target domains, we introduce a global relation matrix that represents the category-wise affinity between domains. 
Specifically, considering the source dataset contains noisy annotations and the target dataset is unlabeled, we first utilize confident proposals after aggregation $\mathbf{P}$ to assemble as batch-wise prototypes, which correspond to the class-wise feature centroids:
\vspace{-8pt}
\begin{equation}
\label{localprototype}
    \{\beta_{(u)}\}_{u=1}^{C} = \frac{1}{Card(\mathbf{P}_{(y')})}\sum_{\substack{y' = u\\
  p_{({y'}, i)} \in \mathbf{P}_{({y})} }} p_{(y',i)},
  \vspace{-7pt}
\end{equation}
where $Card$ is the cardinality and $y'$ is the most confident category. Then, the correspondence between local and global prototypes is characterized according to their semantic correlation, and an adaptive update operation for generating global prototypes $\{\mathcal{B}^{s}_{(u)}\}_{u=1}^{C}$ and $\{\mathcal{B}^{t}_{(v)}\}_{v=1}^{C}$ is conducted:
\vspace{-8pt}
\begin{equation}
    \{\mathcal{B}_{(u)}\}_{u=1}^{C} = \sum_{m=1}^{C} (1-\tau_{(m,u)})\beta_{(m)} + \tau_{(m,u)}\mathcal{B}_{(u)},
\vspace{-7pt}
\end{equation}
where $\tau_{(m,u)}$ is the cosine similarity between the $m$-th batch-wise prototype and the $u$-th global prototype. With this adaptive update process, the representation of global prototypes $\{\mathcal{B}^{s}_{(u)}\}_{u=1}^{C}$ and $\{\mathcal{B}^{t}_{(v)}\}_{v=1}^{C}$ can be strengthened via robust and compact batch-wise local features.
Finally, the global relation matrix $\Pi \in \mathbb{R}^{C\times C}$ is constructed by the cosine similarity between the prototypes, where each entry $\pi_{u,v}$ represents the affinity between the $u$-th prototype in the source domain and the $v$-th prototype in the target domain.

\noindent\textbf{Transition probability regularization.} During alignment, the class-wise domain knowledge of class-corrupted samples are inequilibrium with correctly-labeled samples. To mitigate this, the transition probabilities of the class-corrupted samples are expected to be regularized by the intrinsic class-wise correspondence between source and target domain. Therefore, we directly extract noisy source proposals features from $\Tilde{\mathbf{P}}^s_{(\Tilde{y})}$ regarding to their corresponding noisy labels $\Tilde{y}$, and generate noisy source local prototype similar to the batch-wise prototypes in Eq. (\ref{localprototype}):
\vspace{-8pt}
\begin{equation}
\label{localprototype2}
    \{\Tilde{\beta}^{s}_{(u)}\}_{u=1}^{C} = \frac{1}{Card(\Tilde{\mathbf{P}}^{s}_{(\Tilde{y})})}\sum_{\substack{\Tilde{y} = u\\
  \Tilde{p}^{s}_{(\Tilde{y}, i)} \in \Tilde{\mathbf{P}}^{s}_{(\Tilde{y})} }}\Tilde{p}^{s}_{(\Tilde{y}, i)}.
\vspace{-7pt}
\end{equation}
Then, we build local relation matrix $\mathbf{Z} \in \mathbb{R}^{C\times C}$ between $\Tilde{\beta}^{s}_{(u)}$ and $\mathcal{B}^{t}_{(v)}$ to model the transferability of noisy source samples. Each entry is the class-wise transition probability ${z}_{u, v} = \frac{\Tilde{\beta}^{s}_{(u)} \cdot \mathcal{B}^{t}_{(v)}}{\parallel \Tilde{\beta}^{s}_{(u)} \parallel_2 \cdot \parallel \mathcal{B}^{t}_{(v)} \parallel_2}$. We use $\ell$1 loss to regularize such transition probability between local relation matrix and global relation matrix:
\vspace{-6pt}
\begin{equation}
    \mathcal{L}_{mgrm} = \frac{1}{r}\sum_{r \in \mathds{1}(\mathbf{Z})}|{z}_{r} - {\pi}_{r}|,
\vspace{-6pt}
\end{equation}
where $\mathds{1}(\mathbf{Z})$ refers to the non-zero columns within $\mathbf{Z}$, which indicates the existence of the $r$-th category within the batch. 
Different from other methods that build category-wise graphs \cite{zheng2020cross, ktnet} or maintain batch-wise graphs with fixed shapes \cite{xu2020crossgraph} to model the relationship between source and target domains, our proposed MGRM 
combines the semantic knowledge between source and target domains, and use it to regularize the transition probability of noisy features implicitly. 
Therefore, the transferable representations of feasible adapted noisy samples can be extracted for achieving effective semantic alignment.

\subsection{Entropy-Aware Gradient Reconcilement}

Given noisy annotated source domain data $\mathcal{D}^s$, it implicitly comprises a clean subset $\mathcal{D}^s_{cln} = \{x^s_i, y^s_i\}_{i=1}^{N_{cln}}$ and a subset with both \textit{miss-annotated} and \textit{class-corrupted} samples $\mathcal{D}^s_{cpt} = \{x^s_i, \Tilde{y}^s_i\}_{i=1}^{N_{cpt}}$, which are drawn from the clean and noisy joint distributions $P^{{s}}_{X^{{s}}Y^{{s}}}$ and $\Tilde{P}^{{s}}_{X^{{s}}\Tilde{Y}^{{s}}}$, respectively.
To magnify the effect of learning domain-invariant representations within $\mathcal{D}^s_{cpt}$, we propose an Entropy-Aware Gradient Reconcilement (EAGR) strategy, which first affiliates the class confidence information into the discrimination process, then enforces the gradients of noisy samples to be consistent with the clean ones.

\noindent\textbf{Entropy-aware alignment.} To alleviate the performance deterioration on the target domain, the domain adaptive detector conducts a min-max game to yield a saddle-point solution $\left(\hat{\phi}_{f}, \hat{\phi}_{det}, \hat{\phi}_{dis}\right)$:
\vspace{-8pt}
\begin{equation}
\begin{aligned}
    \left(\hat{\phi}_{f}, \hat{\phi}_{det}\right) &=\arg \min _{{\phi}_{f}, {\phi}_{det}} \mathcal{L}_{det}-\mathcal{L}_{dis}, \\
    \left(\hat{\phi}_{dis}\right) &=\arg \min _{{\phi}_{dis}} \mathcal{L}_{dis},
\end{aligned}
\vspace{-8pt}
\end{equation}
where $\hat{\phi}_{f}$, $\hat{\phi}_{det}$, and $\hat{\phi}_{dis}$ refer to the optimal parameters of the feature extractor, the detector, and the discriminator respectively, which compose the entire domain adaptive detection framework $\hat{\phi}$. However, the noisy labels in the source domain will cause an incompatible optimation between $\left({\phi}_{f}, {\phi}_{det}\right)$ and $\left({\phi}_{dis}\right)$ as the discriminator is class-agnostic, resulting in an insufficient upper boundary of the source risk \cite{CADN, dada}. A natural solution is to directly map category onto features for discrimination \cite{zhao2020adaptive, CADN}, 
but it could further magnify the effect of noisy labels under the noisy DAOD setting if the detector is biased. 
Hence, we build an entropy-aware discriminator to alleviate this effect. Specifically, for each source and target proposal feature $p^s_i \in \mathbf{P}^s, p^t_i \in \mathbf{P}^t$ and their corresponding logits $\eta^s_i \in \boldsymbol{\eta}^s, \eta^t_i \in \boldsymbol{\eta}^t$ generated with $\phi_{det}$, we concatenate them and feed into a discriminator $\phi_{dis}^{EAGR}$, as shown in Fig. \ref{fig:framework}(c). The loss function of the discriminator is written as:
\begin{equation}
\begin{aligned}
    \mathcal{L}_{dis}^{EAGR} = &-\sum_{i,j}zlog(\phi_{dis}^{EAGR}(p^s_i \text{\textcircled{c}} \eta^s_i)) + \\ &(1-z)log(\phi_{dis}^{EAGR}(p^t_j \text{\textcircled{c}} \eta^t_j)),\\
\end{aligned}
\end{equation}
where $z$ refers to the domain label, which is 1 for source and 0 for target. Considering the entropy criterion $H(\eta)=-\sum_{u=1}^{C} \eta_{u} \log (\eta_{u})$ that quantifies
the uncertainty of classifier predictions, the concatenated logits are softly conditioned on the pooled RoI features \cite{cgan0, cgan1, cgan2} to implicitly associate each instance to several
most related categories. Hence, category information is preserved for discriminators in aligning class-wise semantic features within each domain, meanwhile providing entropy-aware gradients for the subsequent gradient concilement process. 
\begin{table*}[htbp]
\setlength\tabcolsep{5pt}
  \centering
  \caption{Results (\%) of Pascal VOC and Noisy Pascal VOC with different noisy rates (NR) $\to$ Clipart1k.}
  \vspace{-5pt}
    \scalebox{0.7}{\begin{tabular}{c|c|cccccccccccccccccccc|cc}
    \toprule
    \multicolumn{23}{c}{Pascal VOC \& Noisy Pascal VOC $\to$ Clipart1k}                        &  \\
    \midrule
    \midrule
    NR    & Methods & aero  & bcycle & bird  & boat  & bottle & bus   & car   & cat   & chair & cow   & table & dog   & hrs   & bike  & prsn  & plnt  & sheep & sofa  & train & tv    & \cellcolor[rgb]{ .816,  .808,  .808}mAP & \cellcolor[rgb]{ .816,  .808,  .808}Imprv. \\
    \midrule
    \multirow{5}[2]{*}{0\%} & DAF   & 29.0  & 45.1  & 33.3  & 25.8  & 28.6  & 48.0  & 39.8  & 12.3  & 35.3  & 50.3  & 22.9  & 17.4  & 33.4  & 33.8  & 59.2  & 44.8  & 20.7  & 26.0  & 45.3  & 49.6  & \cellcolor[rgb]{ .816,  .808,  .808}35.0  & \cellcolor[rgb]{ .816,  .808,  .808}0.0 \\
          & +SCE   & 26.5  & 46.4  & 35.9  & 24.3  & 30.9  & 38.3  & 34.9  & 3.1   & 31.7  & 49.8  & 18.2  & 17.8  & 25.2  & 45.4  & 53.9  & 43.0  & 15.7  & 26.4  & 43.3  & 39.3  & \cellcolor[rgb]{ .816,  .808,  .808}32.5  & \cellcolor[rgb]{ .816,  .808,  .808}-2.5 \\
          & +CP    & 30.3  & 49.2  & 29.8  & 33.2  & 34.1  & 45.8  & 41.1  & 9.7   & 35.8  & 50.7  & 23.6  & 14.4  & 31.7  & 36.9  & 54.6  & 45.8  & 18.6  & 29.9  & 44.8  & 43.5  & \cellcolor[rgb]{ .816,  .808,  .808}35.2  & \cellcolor[rgb]{ .816,  .808,  .808}+0.2 \\
          & +GCE   & 31.9  & 53.2  & 27.9  & 25.8  & 31.0  & 41.9  & 39.3  & 4.3   & 34.5  & 46.7  & 18.1  & 18.4  & 30.2  & 39.1  & 55.0  & 44.1  & 18.1  & 21.1  & 43.2  & 40.7  & \cellcolor[rgb]{ .816,  .808,  .808}33.2  & \cellcolor[rgb]{ .816,  .808,  .808}-1.8 \\
          & +NLTE  & 39.1  & 50.3  & 33.6  & 34.7  & 35.0  & 40.5  & 44.2  & 5.9   & 36.8  & 45.8  & 23.1  & 17.3  & 31.8  & 39.5  & 60.7  & 45.4  & 17.9  & 28.4  & 49.0  & 51.3  & \cellcolor[rgb]{ .816,  .808,  .808}\textbf{36.5} & \cellcolor[rgb]{ .816,  .808,  .808}\textbf{+1.5} \\
    \midrule
    \multirow{5}[2]{*}{20\%} & DAF   & 34.0  & 39.1  & 32.0  & 27.3  & 32.2  & 39.3  & 38.9  & 2.9   & 34.9  & 44.9  & 20.6  & 14.2  & 30.8  & 36.6  & 53.8  & 43.8  & 17.6  & 23.6  & 42.8  & 46.1  & \cellcolor[rgb]{ .816,  .808,  .808}32.8  & \cellcolor[rgb]{ .816,  .808,  .808}0.0 \\
          & +SCE   & 23.3  & 42.3  & 33.1  & 27.3  & 28.8  & 42.4  & 35.1  & 4.0   & 33.0  & 44.2  & 14.6  & 19.4  & 27.0  & 40.9  & 51.1  & 45.2  & 14.9  & 25.8  & 41.6  & 34.8  & \cellcolor[rgb]{ .816,  .808,  .808}31.5  & \cellcolor[rgb]{ .816,  .808,  .808}-1.3 \\
          & +CP    & 29.3  & 39.6  & 29.1  & 28.0  & 29.4  & 34.2  & 42.4  & 3.9   & 35.0  & 39.4  & 21.2  & 12.5  & 32.2  & 38.9  & 57.2  & 43.0  & 18.6  & 27.9  & 40.2  & 45.0  & \cellcolor[rgb]{ .816,  .808,  .808}32.3  & \cellcolor[rgb]{ .816,  .808,  .808}-0.5 \\
          & +GCE   & 24.0  & 42.2  & 32.4  & 29.4  & 31.5  & 45.5  & 39.9  & 6.7   & 36.5  & 38.0  & 16.7  & 15.3  & 30.4  & 37.9  & 53.6  & 44.1  & 13.5  & 24.6  & 46.9  & 43.7  & \cellcolor[rgb]{ .816,  .808,  .808}32.6  & \cellcolor[rgb]{ .816,  .808,  .808}-0.2 \\
          & +NLTE  & 33.1  & 47.5  & 35.5  & 28.2  & 33.7  & 53.8  & 43.8  & 4.2   & 34.2  & 48.4  & 19.3  & 14.6  & 29.7  & 47.2  & 57.1  & 42.5  & 17.7  & 27.7  & 40.0  & 44.5  & \cellcolor[rgb]{ .816,  .808,  .808}\textbf{35.1} & \cellcolor[rgb]{ .816,  .808,  .808}\textbf{+2.3} \\
    \midrule
    \multirow{5}[2]{*}{40\%} & DAF   & 24.5  & 39.4  & 29.1  & 26.9  & 32.8  & 46.5  & 40.0  & 4.7   & 36.1  & 42.0  & 21.3  & 10.6  & 27.8  & 37.3  & 52.8  & 39.7  & 17.5  & 26.9  & 36.0  & 46.2  & \cellcolor[rgb]{ .816,  .808,  .808}31.9  & \cellcolor[rgb]{ .816,  .808,  .808}0.0 \\
          & +SCE   & 17.9  & 42.9  & 29.7  & 21.8  & 26.9  & 41.5  & 34.2  & 8.2   & 29.1  & 38.8  & 19.3  & 19.2  & 28.9  & 48.6  & 50.7  & 42.5  & 10.6  & 20.4  & 41.6  & 40.3  & \cellcolor[rgb]{ .816,  .808,  .808}30.6  & \cellcolor[rgb]{ .816,  .808,  .808}-1.3 \\
          & +CP    & 24.0  & 40.9  & 31.1  & 22.0  & 31.2  & 33.0  & 40.8  & 4.4   & 34.6  & 36.8  & 18.5  & 13.8  & 29.6  & 41.3  & 51.7  & 38.6  & 14.2  & 27.8  & 26.0  & 37.8  & \cellcolor[rgb]{ .816,  .808,  .808}29.9  & \cellcolor[rgb]{ .816,  .808,  .808}-2.0 \\
          & +GCE   & 27.0  & 36.2  & 31.1  & 26.0  & 33.8  & 42.9  & 41.2  & 2.6   & 37.0  & 45.3  & 19.0  & 17.2  & 33.0  & 42.9  & 54.6  & 45.7  & 17.6  & 21.4  & 41.9  & 49.1  & \cellcolor[rgb]{ .816,  .808,  .808}33.3  & \cellcolor[rgb]{ .816,  .808,  .808}+1.4 \\
          & +NLTE  & 32.8  & 45.5  & 30.8  & 29.8  & 35.7  & 43.2  & 43.0  & 6.4   & 32.7  & 45.9  & 19.8  & 10.8  & 31.1  & 43.4  & 56.4  & 43.3  & 19.6  & 24.8  & 42.5  & 43.9  & \cellcolor[rgb]{ .816,  .808,  .808}\textbf{34.1} & \cellcolor[rgb]{ .816,  .808,  .808}\textbf{+2.2} \\
    \midrule
    \multirow{5}[2]{*}{60\%} & DAF   & 29.4  & 33.5  & 29.7  & 29.0  & 27.7  & 39.5  & 38.0  & 2.7   & 31.9  & 41.5  & 19.8  & 12.9  & 30.2  & 37.0  & 49.7  & 37.2  & 12.8  & 25.5  & 40.8  & 44.2  & \cellcolor[rgb]{ .816,  .808,  .808}30.6  & \cellcolor[rgb]{ .816,  .808,  .808}0.0 \\
          & +SCE   & 22.2  & 44.0  & 31.3  & 28.0  & 29.8  & 48.7  & 31.6  & 11.0  & 29.1  & 30.7  & 19.7  & 9.2   & 25.8  & 55.9  & 51.9  & 41.3  & 5.7   & 21.8  & 49.0  & 34.9  & \cellcolor[rgb]{ .816,  .808,  .808}31.1  & \cellcolor[rgb]{ .816,  .808,  .808}+0.5 \\
          & +CP    & 32.2  & 42.1  & 31.5  & 26.3  & 31.9  & 42.4  & 40.5  & 2.7   & 31.8  & 45.2  & 20.0  & 12.2  & 26.5  & 38.1  & 51.1  & 42.3  & 11.0  & 25.6  & 38.4  & 41.4  & \cellcolor[rgb]{ .816,  .808,  .808}31.7  & \cellcolor[rgb]{ .816,  .808,  .808}+1.1 \\
          & +GCE   & 26.7  & 43.3  & 33.0  & 28.6  & 33.8  & 51.8  & 38.0  & 6.3   & 33.8  & 41.7  & 22.2  & 13.9  & 33.4  & 44.9  & 53.1  & 43.9  & 14.5  & 22.8  & 38.6  & 43.7  & \cellcolor[rgb]{ .816,  .808,  .808}33.3  & \cellcolor[rgb]{ .816,  .808,  .808}+2.7 \\
          & +NLTE  & 33.0  & 51.9  & 32.2  & 31.7  & 29.9  & 39.7  & 43.6  & 11.0  & 36.4  & 40.7  & 27.0  & 11.8  & 30.3  & 35.3  & 55.9  & 42.2  & 20.8  & 30.1  & 34.5  & 41.2  & \cellcolor[rgb]{ .816,  .808,  .808}\textbf{34.0} & \cellcolor[rgb]{ .816,  .808,  .808}\textbf{+3.4} \\
    \midrule
    \multirow{5}[2]{*}{80\%} & DAF   & 28.2  & 34.0  & 29.6  & 20.8  & 27.7  & 45.0  & 34.4  & 1.4   & 31.5  & 34.1  & 19.9  & 9.3   & 26.2  & 33.3  & 46.0  & 37.4  & 17.5  & 20.4  & 30.6  & 41.9  & \cellcolor[rgb]{ .816,  .808,  .808}28.5  & \cellcolor[rgb]{ .816,  .808,  .808}0.0 \\
          & +SCE   & 19.5  & 32.9  & 28.9  & 23.1  & 34.3  & 50.6  & 31.5  & 4.3   & 29.5  & 35.9  & 19.5  & 12.6  & 23.9  & 56.2  & 52.6  & 38.0  & 8.2   & 21.7  & 41.8  & 35.5  & \cellcolor[rgb]{ .816,  .808,  .808}30.0  & \cellcolor[rgb]{ .816,  .808,  .808}+1.5 \\
          & +CP    & 25.2  & 36.1  & 27.5  & 29.8  & 32.5  & 29.1  & 34.3  & 3.2   & 31.4  & 37.7  & 22.3  & 7.6   & 30.4  & 36.5  & 46.8  & 35.4  & 19.9  & 27.0  & 29.6  & 39.1  & \cellcolor[rgb]{ .816,  .808,  .808}29.1  & \cellcolor[rgb]{ .816,  .808,  .808}+0.6 \\
          & +GCE   & 25.8  & 32.8  & 29.2  & 21.1  & 28.8  & 50.0  & 33.5  & 1.3   & 28.9  & 34.1  & 21.7  & 7.7   & 27.2  & 46.8  & 50.1  & 37.9  & 7.3   & 20.2  & 42.5  & 36.0  & \cellcolor[rgb]{ .816,  .808,  .808}29.1  & \cellcolor[rgb]{ .816,  .808,  .808}+0.6 \\
          & +NLTE  & 36.0  & 45.4  & 33.5  & 30.3  & 27.3  & 40.5  & 40.6  & 2.6   & 28.3  & 51.7  & 20.4  & 9.5   & 30.8  & 43.1  & 56.6  & 42.1  & 17.7  & 23.3  & 31.2  & 38.4  & \cellcolor[rgb]{ .816,  .808,  .808}\textbf{32.5} & \cellcolor[rgb]{ .816,  .808,  .808}\textbf{+4.0} \\
    \bottomrule
    \bottomrule
    \end{tabular}}%
  \label{tab:clipart}%
  \vspace{-12pt}
\end{table*}%

\noindent\textbf{Gradient reconcilement.} 
Given a domain adaptive detector with parameters $\phi$ and objective function $\mathcal{L}$, we have gradients for different roles of the proposals:
\vspace{-6pt}
\begin{equation}
\begin{aligned}
    G^{s}_{cln} &= \mathbb{E}_{x^s \in D^s_{cln}} \frac{\partial\mathcal{L}[(x^s, y^s) ; \phi]}{\partial \phi},\\
    G^{s}_{cpt} &= \mathbb{E}_{x^s \in D^s_{cpt}} \frac{\partial\mathcal{L}[(x^s, \Tilde{y}^s) ; \phi]}{\partial \phi},\\
    G^{t} &= \mathbb{E}_{x^t \in D^t} \frac{\partial\mathcal{L}[(x^t) ; \phi]}{\partial \phi},
\end{aligned}
\vspace{-6pt}
\end{equation}
where $G^{s}_{cln}$, $G^{s}_{cpt}$, $G^{t}$ are the gradients provided by clean proposals, noisy proposals, and target proposals, respectively. 
%
After the entropy-aware alignment, the gradients $G^{s}_{cln}$, $G^{s}_{cpt}$, and $G^{t}$ are conditioned to the class-wise information provided by both noisy labels and the entropy of classifier predictions. 
Considering that both $G^{s}_{cln}$ and $G^{t}$ optimize the feature extractor $\hat{\phi}_{f}$ and detector $\hat{\phi}_{det}$ in the direction towards learning domain-invariant representations, the value of their inner product $G^{s}_{cln} \cdot G^{t}$ is expected to be sufficiently large. 
Nevertheless, the direction of $G^{s}_{cpt}$ could not be determined as they are produced by noisy samples. 
To reliably characterize the domain-invariant portion within $G^{s}_{cpt}$, we simultaneously maximize $G^{s}_{cpt} \cdot G^{s}_{cln}$ and $G^{s}_{cpt} \cdot G^{t}$ to encourage the direction of gradients provided by noisy and clean samples to be consistent. 
Thus, we are expected to maximize the summation of the above terms:
\vspace{-4pt}
\begin{equation}
     \arg \max _{{\phi}_{f}, {\phi}_{det}, {\phi}_{dis}}  \big(G^{s}_{cln} \cdot G^{s}_{cpt} + G^{s}_{cln} \cdot G^{t} + G^{s}_{cpt} \cdot G^{t}\big).
\vspace{-4pt}
\label{argmaxgrad}
\end{equation}
As we cannot directly optimize Eq. (\ref{argmaxgrad}) with SGD as clean and noisy gradients cannot be explicitly split, meanwhile computing the Hessians (second order derivatives) is computational prohibitive, inspired by \cite{reptile, fish}, we utilize the first-order meta update of the network as an approximation,
which could maximize the above inner product between gradients over iterations and avoid splitting $G^{s}_{cln}$ and $G^{s}_{cpt}$ :
\begin{equation}
\label{eagr_meta}
    ({\phi}_{f}, {\phi}_{det}) \leftarrow ({\phi}_{f}, {\phi}_{det}) + \lambda(\Delta\Tilde{\phi}_{f}, \Delta\Tilde{\phi}_{det}),
\end{equation}
where $(\Delta\Tilde{\phi}_{f}, \Delta\Tilde{\phi}_{det})$ denotes the residual of the parameters before and after multi-step training of the network and $\lambda$ is the meta weight.
The detailed proof of the approximation is provided in the supplementary.
With EAGR, the semantic and discrimination information are harmonized into the backpropagation process, and the gradients of distinct samples are encouraged to achieve coherence. Therefore, both clean and noisy samples would be contributive towards learning a domain-invariant object detector.

\subsection{Framework Optimization}
The framework is trained with the following objective function:
\begin{equation}
\label{finalloss}
    \mathcal{L}=\mathcal{L}_{det}+\lambda_{mgrm}\mathcal{L}_{mgrm}+\mathcal{L}_{dis}^{DAF}+\mathcal{L}_{dis}^{EAGR},
\end{equation}
where $\mathcal{L}_{det}$ denotes the loss of Faster R-CNN \cite{fasterrcnn} which consists of RPN loss and RCNN loss. $\mathcal{L}_{dis}^{DAF}$ contains the discrimination components in DAF \cite{DAfasterrcnn}. Hereafter, we adopt meta update in Eq. (\ref{eagr_meta}) for achieving gradient reconcilement.
During inference, the input images are consecutively fed into $({\phi}_{f}, {\phi}_{det})$ to obtain the detection results.

\section{Experiments}
In this section, we first introduce the experimental setup of synthetic noise and real-world noise, then compare NLTE with the baseline DAF \cite{DAfasterrcnn} and equipping different noise-robust training approaches \cite{cp,sce,gce}. Also, NLTE is compared with existing DAOD methods \cite{DAfasterrcnn, scda, mtor, xu2020crossgraph, everypixelmatters,zhao2020adaptive, zhang2021rpn, dss, DIDN, wu2021vector, STAL} on the real-world noise setting. 

\subsection{Experimental Setup}

\subsubsection{Datasets}

\noindent\textbf{Pascal VOC \& Noisy Pascal VOC.} Pascal VOC \cite{everingham2015pascal} contains
16,551 images with 20 distinct object categories. As it contains few instances per image and has been extensively verified by human annotators, we consider it as a clean dataset with no noise. Based on the clean Pascal VOC, we randomly add synthetic label noise with different rates to mimic the annotation mistakes. Specifically, we randomly select a portion of samples and substitute them to another random label. Note that if a label is substituted to background, the corresponding instance is directly removed.

\noindent\textbf{Clipart1k \& Watercolor2k.} Clipart1k \cite{inoue2018cross} contains 1k graphical images and shares the same 20 categories as Pascal VOC. All images are used for both adversarial training and testing. Watercolor2k \cite{inoue2018cross} shares 6 common categories as the Pascal VOC dataset. We use the 1k training set for adversarial training and the remaining 1k for testing.

\noindent\textbf{Cityscapes \& Foggy Cityscapes.} Cityscapes \cite{cordts2016cityscapes} contains 2,975 images for training and 500 images for validation. As shown in Fig. \ref{fig:teaser}, Cityscapes dataset contains noisy annotations itself, so we treat it as noisy annotated and directly use the training set as the source domain. Foggy Cityscapes \cite{sakaridis2018foggy} is a fog-rendered Cityscapes dataset and we follow \cite{DAfasterrcnn, xu2020crossgraph, sapn, everypixelmatters} to use the validation set as the target domain. As the validation set only contains 500 images, we manually check all images and consider it as a clean dataset.
\vspace{-20pt}
\subsubsection{Training Details}
For all experiments, DA Faster R-CNN (DAF) \cite{DAfasterrcnn} with backbone ResNet-50 \cite{resnet} is utilized as our baseline UDA object detector. SGD optimizer is used for training the model for 7 epochs, with an initial learning rate $1\times10^{-3}$ and decays by 0.1 after 5 and 6 epochs. Following the common practice \cite{swda, sapn}, we resize the shorter side of the image to 600 during both training and testing unless specified. $\lambda_{mgrm}$ is set to 0.1. Experiments are conducted on NVIDIA V100 GPUs and PyTorch is used for the implementation.
\begin{table}[t]
  \centering
  \caption{Results (\%) of Pascal VOC and Noisy Pascal VOC with different noisy rates (NR) $\to$ Watercolor2k.}
  \vspace{-8pt}
    \scalebox{0.7}{\begin{tabular}{c|c|cccccc|cc}
    \toprule
    \multicolumn{9}{c}{\qquad\qquad\qquad Pascal VOC \& Noisy Pascal VOC $\to$ Watercolor2k}          &  \\
    \midrule
    \midrule
    NR    & Methods & bcycle & bird  & car   & cat   & dog   & prsn  & \cellcolor[rgb]{ .816,  .808,  .808}mAP & \cellcolor[rgb]{ .816,  .808,  .808}Imprv. \\
    \midrule
    \multirow{5}[2]{*}{0\%} & DAF   & 65.8  & 40.4  & 35.3  & 30.0  & 21.5  & 44.1  & \cellcolor[rgb]{ .816,  .808,  .808}39.6 & \cellcolor[rgb]{ .816,  .808,  .808}0.0 \\
          & +SCE   & 65.3  & 36.9  & 38.3  & 25.8  & 18.9  & 43.2  & \cellcolor[rgb]{ .816,  .808,  .808}37.9 & \cellcolor[rgb]{ .816,  .808,  .808}-1.7 \\
          & +CP    & 67.1  & 39.1  & 34.5  & 27.2  & 22.9  & 45.3  & \cellcolor[rgb]{ .816,  .808,  .808}39.4 & \cellcolor[rgb]{ .816,  .808,  .808}-0.2 \\
          & +GCE   & 67.3  & 37.0  & 39.7  & 21.9  & 21.3  & 46.4  & \cellcolor[rgb]{ .816,  .808,  .808}38.9 & \cellcolor[rgb]{ .816,  .808,  .808}-0.7 \\
          & +NLTE  & 73.7  & 36.9  & 39.9  & 26.8  & 22.6  & 45.3  & \cellcolor[rgb]{ .816,  .808,  .808}\textbf{40.9} & \cellcolor[rgb]{ .816,  .808,  .808}\textbf{+1.3} \\
    \midrule
    \multirow{5}[2]{*}{20\%} & DAF   & 69.1  & 36.5  & 25.8  & 31.0  & 16.1  & 44.9  & \cellcolor[rgb]{ .816,  .808,  .808}37.2 & \cellcolor[rgb]{ .816,  .808,  .808}0.0 \\
          & +SCE   & 62.4  & 42.6  & 33.2  & 32.2  & 18.5  & 46.5  & \cellcolor[rgb]{ .816,  .808,  .808}39.2 & \cellcolor[rgb]{ .816,  .808,  .808}+2.0 \\
          & +CP    & 72    & 36.5  & 21.3  & 18.3  & 21.1  & 41.5  & \cellcolor[rgb]{ .816,  .808,  .808}35.1 & \cellcolor[rgb]{ .816,  .808,  .808}-2.1 \\
          & +GCE   & 62.7  & 42.5  & 40.1  & 26.2  & 18.8  & 44.9  & \cellcolor[rgb]{ .816,  .808,  .808}39.2 & \cellcolor[rgb]{ .816,  .808,  .808}+2.0 \\
          & +NLTE  & 73.7  & 37.1  & 35.3  & 28.1  & 21.2  & 44.5  & \cellcolor[rgb]{ .816,  .808,  .808}\textbf{40.0} & \cellcolor[rgb]{ .816,  .808,  .808}\textbf{+2.8} \\
    \midrule
    \multirow{5}[2]{*}{40\%} & DAF   & 68.0  & 32.9  & 20.5  & 19.8  & 13.6  & 39.4  & \cellcolor[rgb]{ .816,  .808,  .808}32.4 & \cellcolor[rgb]{ .816,  .808,  .808}0.0 \\
          & +SCE   & 64.5  & 36.6  & 37.8  & 14.1  & 14.0  & 42.8  & \cellcolor[rgb]{ .816,  .808,  .808}35.0 & \cellcolor[rgb]{ .816,  .808,  .808}+2.6 \\
          & +CP    & 66.0  & 36.6  & 17.8  & 24.0  & 18.2  & 39.8  & \cellcolor[rgb]{ .816,  .808,  .808}33.7 & \cellcolor[rgb]{ .816,  .808,  .808}+1.3 \\
          & +GCE   & 64.3  & 40.0  & 34.7  & 21.3  & 19.0  & 43.8  & \cellcolor[rgb]{ .816,  .808,  .808}37.2 & \cellcolor[rgb]{ .816,  .808,  .808}+4.8 \\
          & +NLTE  & 75.7  & 37.2  & 32.5  & 22.6  & 24.3  & 43.1  & \cellcolor[rgb]{ .816,  .808,  .808}\textbf{39.2} & \cellcolor[rgb]{ .816,  .808,  .808}\textbf{+6.8} \\
    \midrule
    \multirow{5}[2]{*}{60\%} & DAF   & 58.6  & 35.6  & 16.7  & 18.8  & 11.5  & 40.1  & \cellcolor[rgb]{ .816,  .808,  .808}30.2 & \cellcolor[rgb]{ .816,  .808,  .808}0.0 \\
          & +SCE   & 68.1  & 36.3  & 31.8  & 21.9  & 19.7  & 41.3  & \cellcolor[rgb]{ .816,  .808,  .808}36.5 & \cellcolor[rgb]{ .816,  .808,  .808}+6.3 \\
          & +CP    & 68.4  & 30.3  & 24.0  & 22.8  & 9.6   & 38.7  & \cellcolor[rgb]{ .816,  .808,  .808}32.3 & \cellcolor[rgb]{ .816,  .808,  .808}+2.1 \\
          & +GCE   & 73.7  & 33.0  & 28.7  & 24.3  & 20.4  & 41.2  & \cellcolor[rgb]{ .816,  .808,  .808}36.9 & \cellcolor[rgb]{ .816,  .808,  .808}+6.7 \\
          & +NLTE  & 69.5  & 35.4  & 27.4  & 28.4  & 19.8  & 51.5  & \cellcolor[rgb]{ .816,  .808,  .808}\textbf{38.6} & \cellcolor[rgb]{ .816,  .808,  .808}\textbf{+8.4} \\
    \midrule
    \multirow{5}[2]{*}{80\%} & DAF   & 56.8  & 36.7  & 15.6  & 19.0    & 14.8  & 37.8  & \cellcolor[rgb]{ .816,  .808,  .808}30.1 & \cellcolor[rgb]{ .816,  .808,  .808}0.0 \\
          & +SCE   & 69.4  & 37.4  & 22.6  & 24.3  & 16.6  & 34.6  & \cellcolor[rgb]{ .816,  .808,  .808}34.2 & \cellcolor[rgb]{ .816,  .808,  .808}+4.1 \\
          & +CP    & 49.1  & 36.1  & 16.6  & 13.7  & 10.1  & 36.9  & \cellcolor[rgb]{ .816,  .808,  .808}27.1 & \cellcolor[rgb]{ .816,  .808,  .808}-3.0 \\
          & +GCE   & 62.8  & 34.3  & 14.5  & 13.4  & 10.7  & 40.6  & \cellcolor[rgb]{ .816,  .808,  .808}29.4 & \cellcolor[rgb]{ .816,  .808,  .808}-0.7 \\
          & +NLTE  & 72.7  & 41.4  & 6.6   & 30.5  & 14.1  & 47.9  & \cellcolor[rgb]{ .816,  .808,  .808}\textbf{35.6} & \cellcolor[rgb]{ .816,  .808,  .808}\textbf{+5.5} \\
    \bottomrule
    \bottomrule
    \end{tabular}}%
  \label{tab:p2w}%
  \vspace{-12pt}
\end{table}%

\vspace{-2pt}
\subsection{Synthetic Noise}
\noindent\textbf{Pascal VOC \& Noisy Pascal VOC $\to$ Clipart1k}.
We list the results of using Pascal VOC and Noisy Pascal VOC with different noisy rates as the source domain, and Clipart1k as the target domain in Table \ref{tab:clipart}. It is shown that the performance of the baseline domain adaptive detector \cite{DAfasterrcnn} drops consistently as the noisy rate increases, and drops from $35.0\%$ to $28.5\%$ under $80\%$ noisy annotations. With loss adjustment method CP \cite{cp}, the performance shows limited improvement and even drops by $0.5\%$ and $2.0\%$ at $20\%$ and $40\%$ noisy rates. With symmetric loss methods SCE \cite{sce} and GCE \cite{gce}, the detector performs better than CP under noisy settings, but they significantly deteriorate the performance of the detector under clean scenario, causing the mAP drops by $2.5\%$ and $1.8\%$, respectively. However, adding our proposed NLTE not only achieves robust adaptive detection under different noisy rates ($2.3\%$ mAP improvement at $20\%$ and $4.0\%$ mAP improvement at $80\%$), but also guarantees the performance of the domain adaptive detector when the source annotations are clean.

\begin{table}[t]
\setlength\tabcolsep{4.5pt}
  \centering
  \caption{Results (\%) of Cityscapes $\to$ Foggy Cityscapes. $\dagger$ denotes larger training and testing scales.}
  \vspace{-8pt}
    \scalebox{0.67}{\begin{tabular}{c|cccccccc|c}
    \toprule
    \multicolumn{10}{c}{Cityscapes $\to$ Foggy Cityscapes} \\
    \midrule
    \midrule
    Methods & prsn  & rider & car   & truck & bus   & train & mtor  & bike  & \cellcolor[rgb]{ .816,  .808,  .808}mAP \\
    \midrule
    Source-only\cite{fasterrcnn} & 26.9  & 38.2  & 35.6  & 18.3  & 32.4  & 9.6   & 25.8  & 28.6  & \cellcolor[rgb]{ .816,  .808,  .808}26.9 \\
    DAF\cite{DAfasterrcnn}$_{CVPR'18}$   & 25.0  & 31.0  & 40.5  & 22.1  & 35.3  & 20.2  & 20.0  & 27.1  & \cellcolor[rgb]{ .816,  .808,  .808}27.6 \\
    SC-DA\cite{scda}$_{CVPR'19}$ & 33.5  & 38.0  & 48.5  & 26.5  & 39.0  & 23.3  & 28.0  & 33.6  & \cellcolor[rgb]{ .816,  .808,  .808}33.8 \\
    MTOR\cite{mtor}$_{CVPR'19}$  & 30.6  & 41.4  & 44.0  & 21.9  & 38.6  & 40.6  & 28.3  & 35.6  & \cellcolor[rgb]{ .816,  .808,  .808}35.1 \\
    GPA\cite{xu2020crossgraph}$_{CVPR'20}$   & 32.9  & 46.7  & 54.1  & 24.7  & 45.7  & 41.1  & 32.4  & 38.7  & \cellcolor[rgb]{ .816,  .808,  .808}39.5 \\
    MCAR \cite{zhao2020adaptive}$_{ECCV'20}$& 32.0&42.1&43.9&31.3&44.1&43.4&37.4&36.6&\cellcolor[rgb]{ .816,  .808,  .808}38.8\\
    EPM$^\dagger$\cite{everypixelmatters}$_{ECCV'20}$  & 41.9  & 38.7  & 56.7  & 22.6  & 41.5  & 26.8  & 24.6  & 35.5  & \cellcolor[rgb]{ .816,  .808,  .808}36.0 \\
    RPNPA\cite{zhang2021rpn}$_{CVPR'21}$ & 33.3  & 45.6  & 50.5  & 30.4  & 43.6  & {42.0} & 29.7  & 36.8  & \cellcolor[rgb]{ .816,  .808,  .808}39.0 \\
    DSS-UDA\cite{dss}$_{CVPR'21}$ & 42.9  & {51.2} & 53.6  & {33.6} & 49.2  & 18.9  & {36.2} & 41.8  & \cellcolor[rgb]{ .816,  .808,  .808}40.9 \\
    DIDN\cite{DIDN}$_{ICCV'21}$  & 38.3  & 44.4  & 51.8  & 28.7  & 53.3  & 34.7  & 32.4  & 40.4  & \cellcolor[rgb]{ .816,  .808,  .808}40.5 \\
    VDD\cite{wu2021vector}$_{ICCV'21}$   & 33.4  & 44.0  & 51.7  & {33.9} & {52.0} & 40.9  & 32.3  & 36.8  & \cellcolor[rgb]{ .816,  .808,  .808}{39.8} \\
    SSAL$^\dagger$\cite{STAL}$_{NeurIPS'21}$ & {45.1} & 47.4  & {59.4} & 24.5  & 50.0    & 25.7  & 26.0  & 38.7  & \cellcolor[rgb]{ .816,  .808,  .808}39.6 \\
    \midrule
    \textbf{\textit{DAF+NLTE (Ours)}} & 37.0  & 46.9  & 54.8  & 32.1  & 49.9  & 43.5  & 29.9  & 39.6  & \cellcolor[rgb]{ .816,  .808,  .808}41.8 \\
    \textbf{\textit{DAF+NLTE (Ours)$^\dagger$}} & {43.1} & 50.7  & {58.7} & {33.6} & {56.7} & {42.7} & 33.7  & {43.3}  & \cellcolor[rgb]{ .816,  .808,  .808}\textbf{45.4} \\
    \bottomrule
    \bottomrule
    \end{tabular}}%
  \label{tab:cs}%
  \vspace{-18pt}
\end{table}%
\begin{figure*}[t]
	\centering
	\footnotesize
   \begin{tabular}{c@{\hskip2pt}c@{\hskip2pt}c@{\hskip2pt}c@{\hskip2pt}c}
   \includegraphics[width=3.0cm]{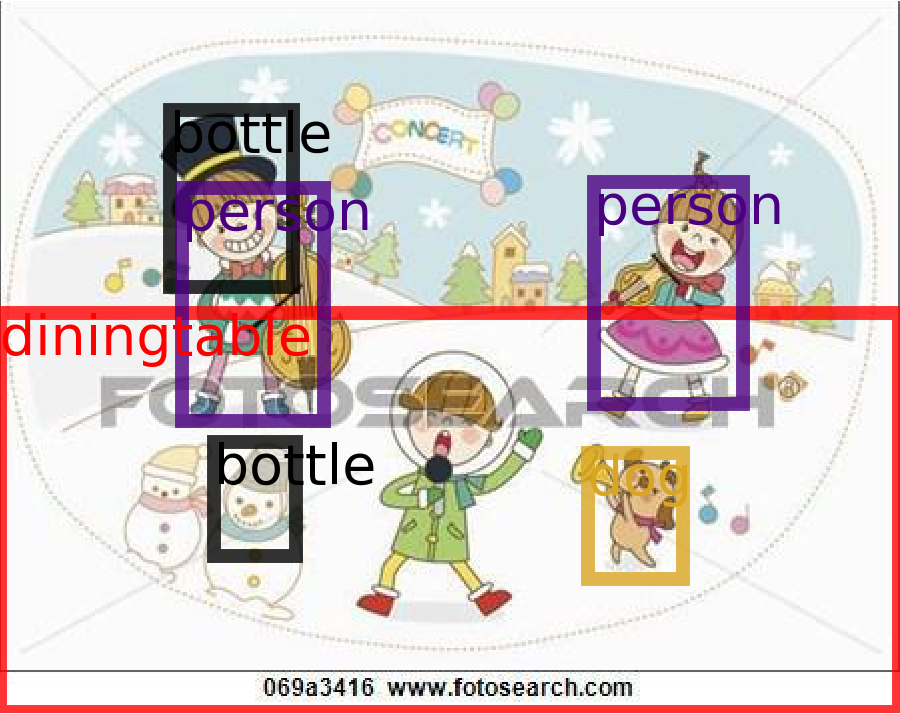}&%
		\includegraphics[width=3.0cm]{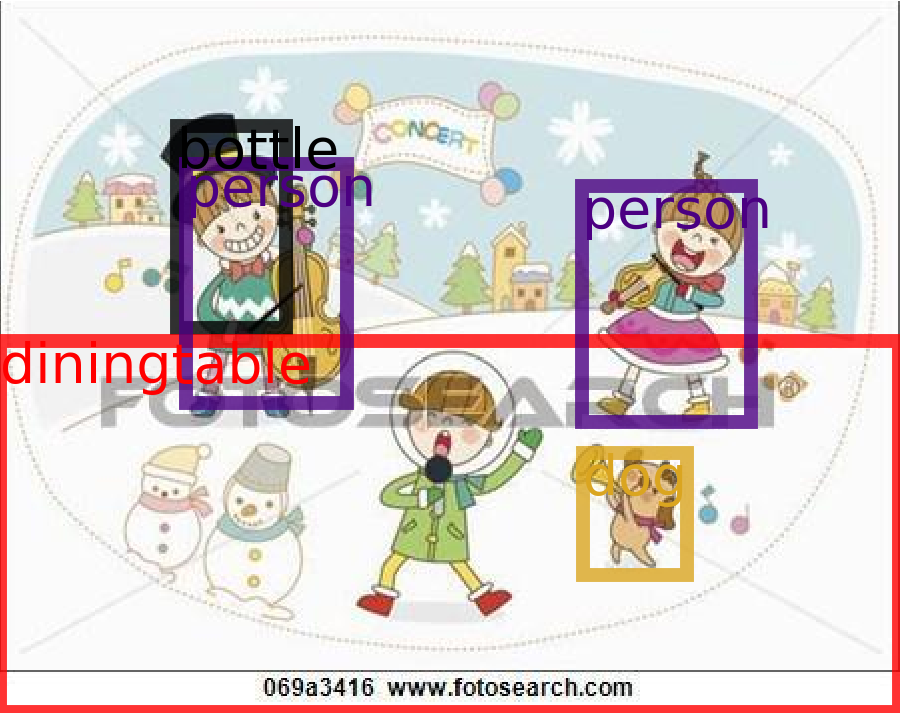}&%
		\includegraphics[width=3.0cm]{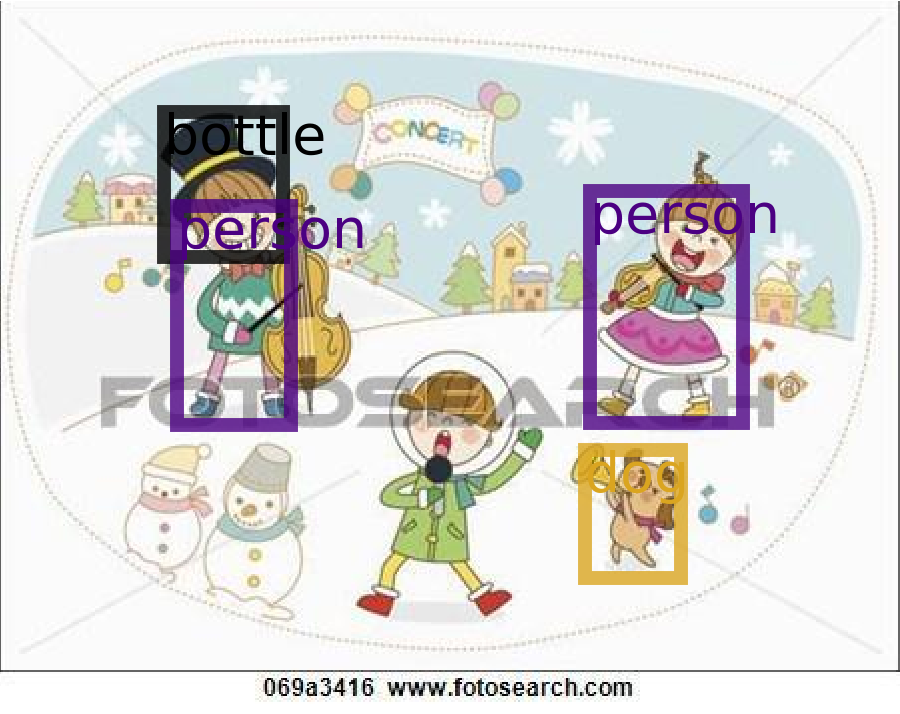}&%
      \includegraphics[width=3.0cm]{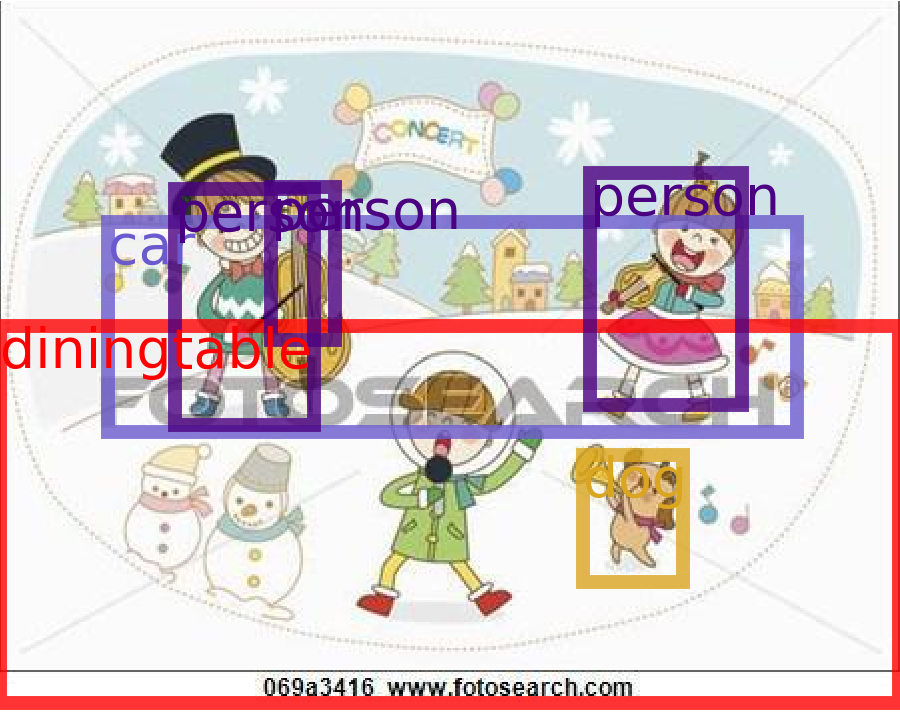}&
      \includegraphics[width=3.0cm]{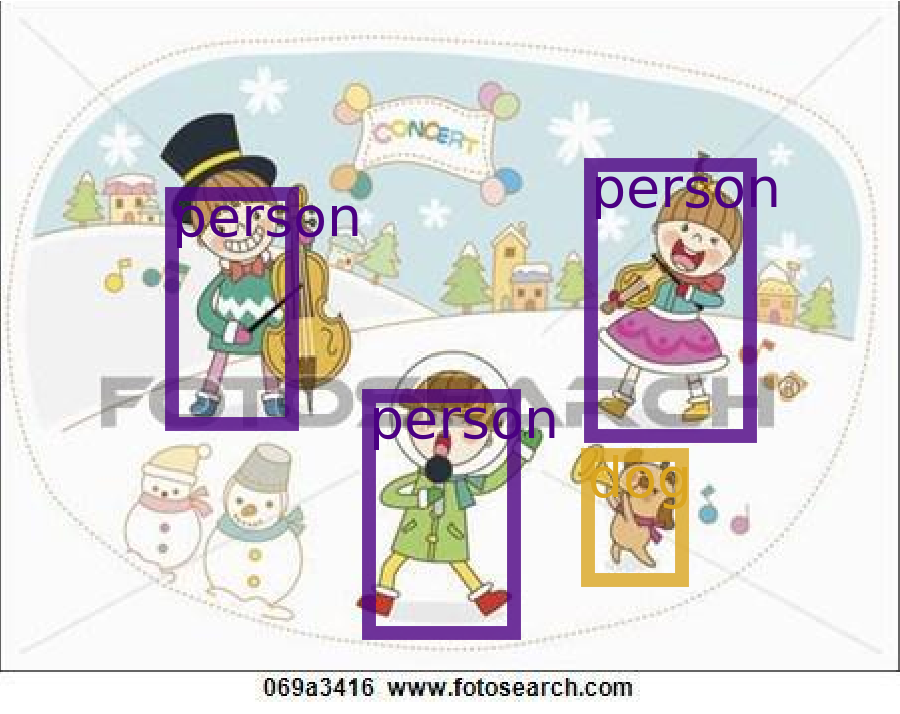}\\
      \includegraphics[width=3.0cm]{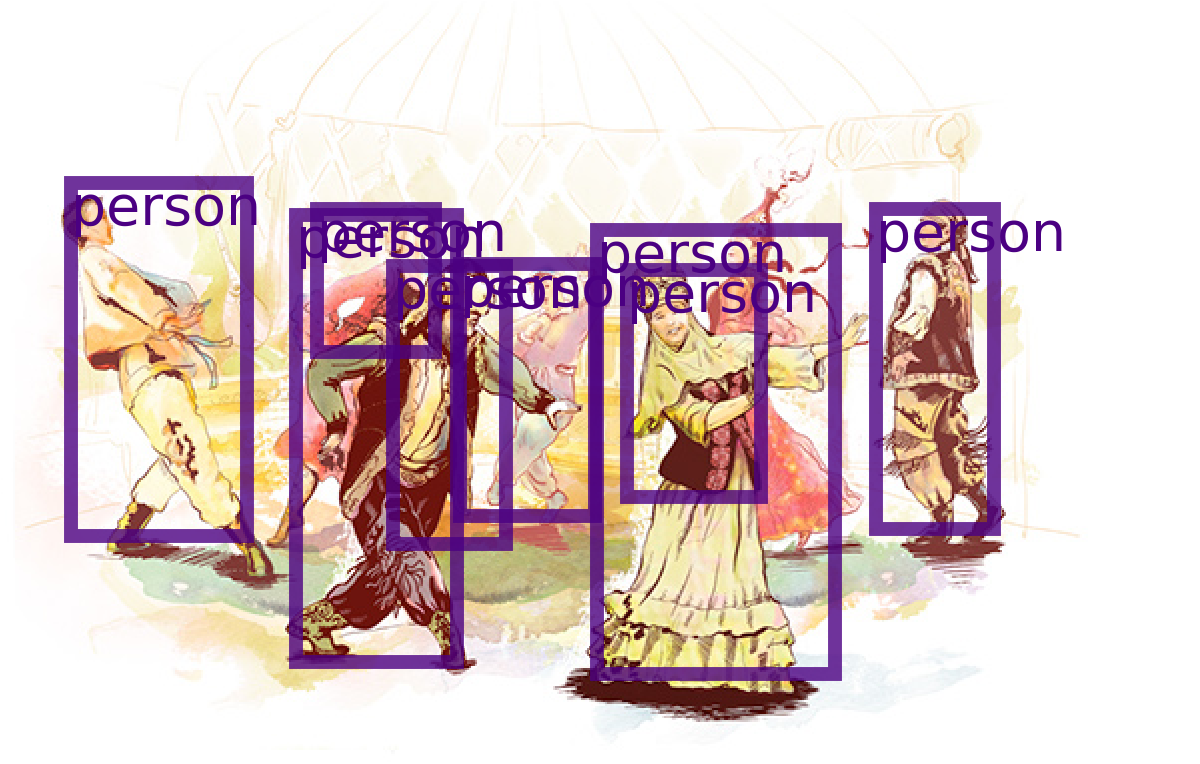}&%
		\includegraphics[width=3.0cm]{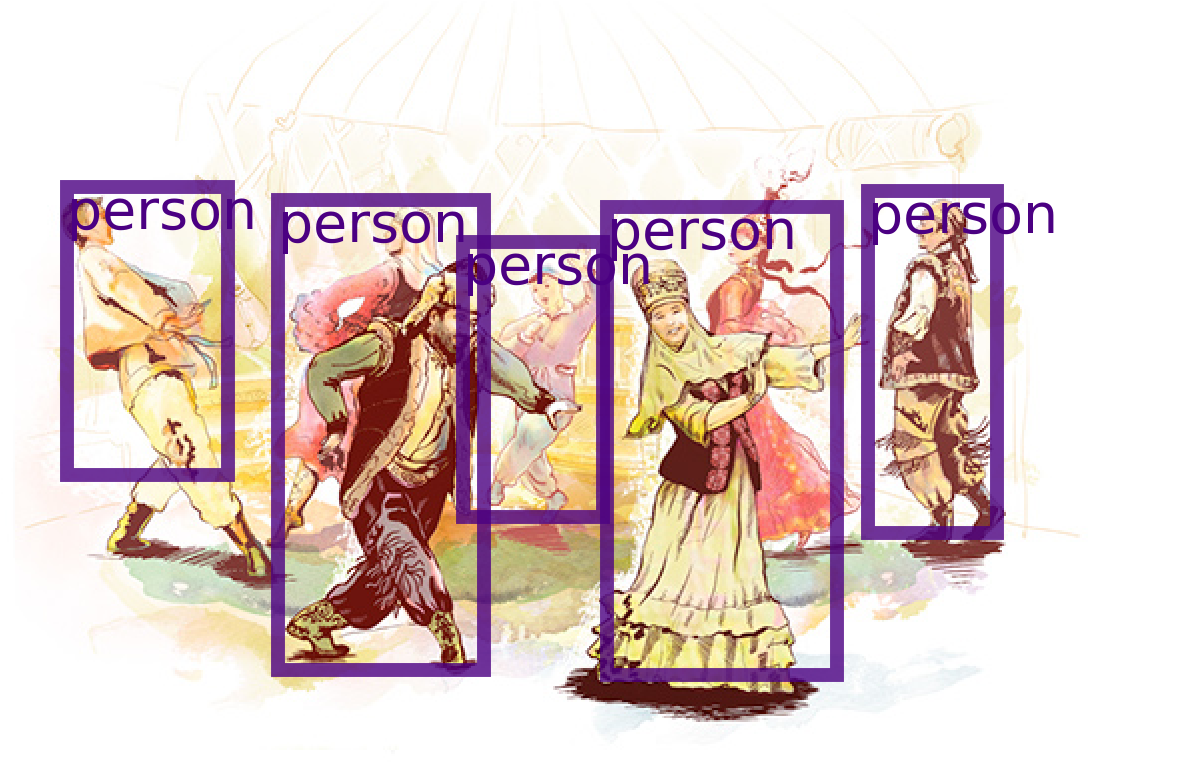}&%
		\includegraphics[width=3.0cm]{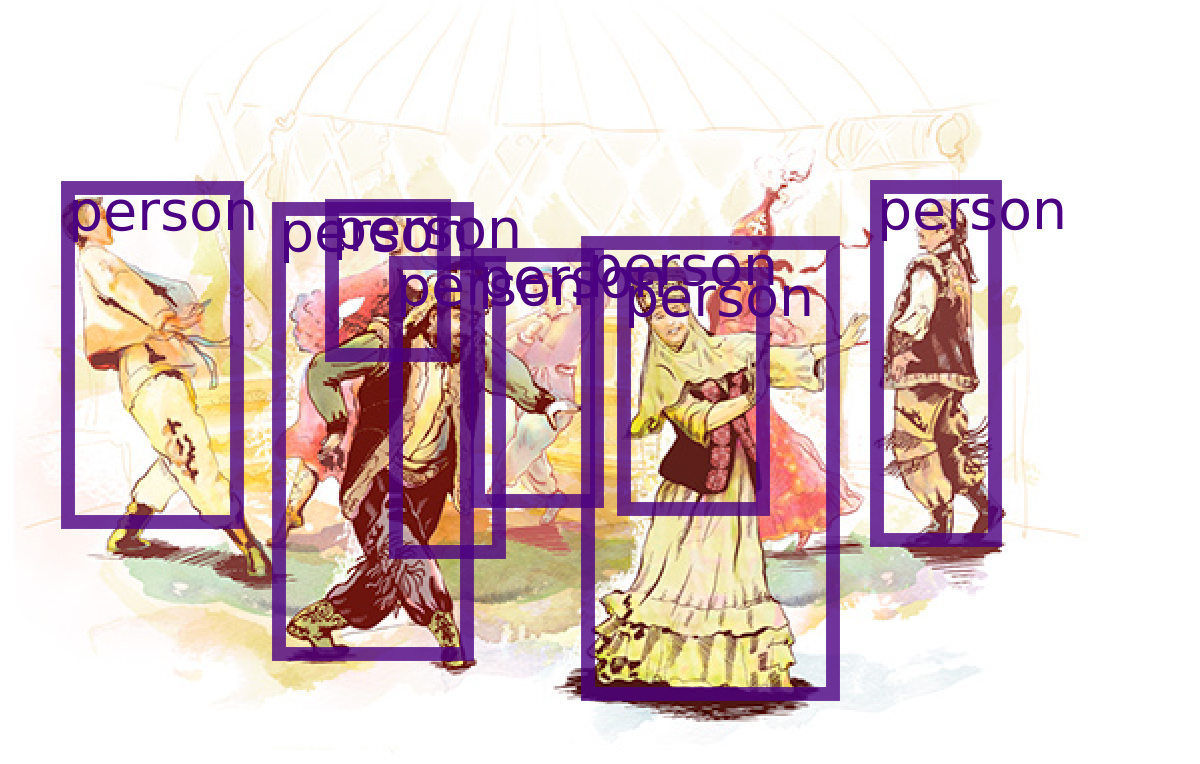}&%
      \includegraphics[width=3.0cm]{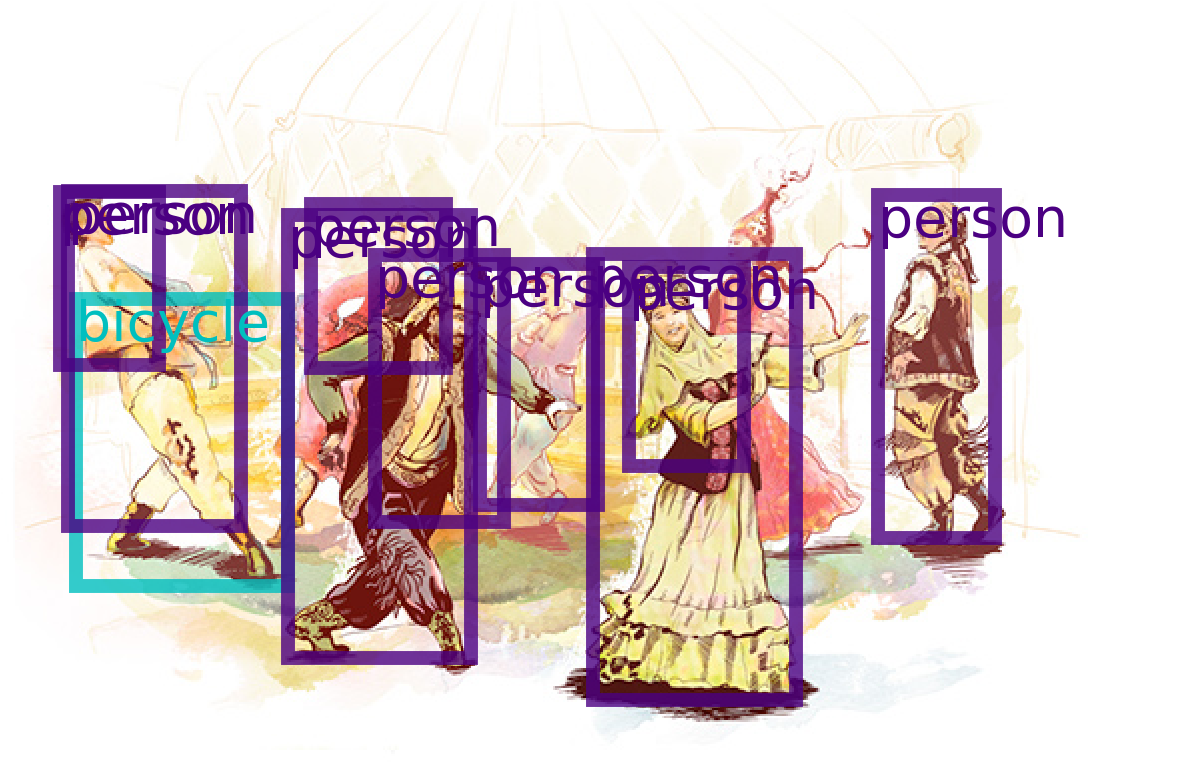}&
      \includegraphics[width=3.0cm]{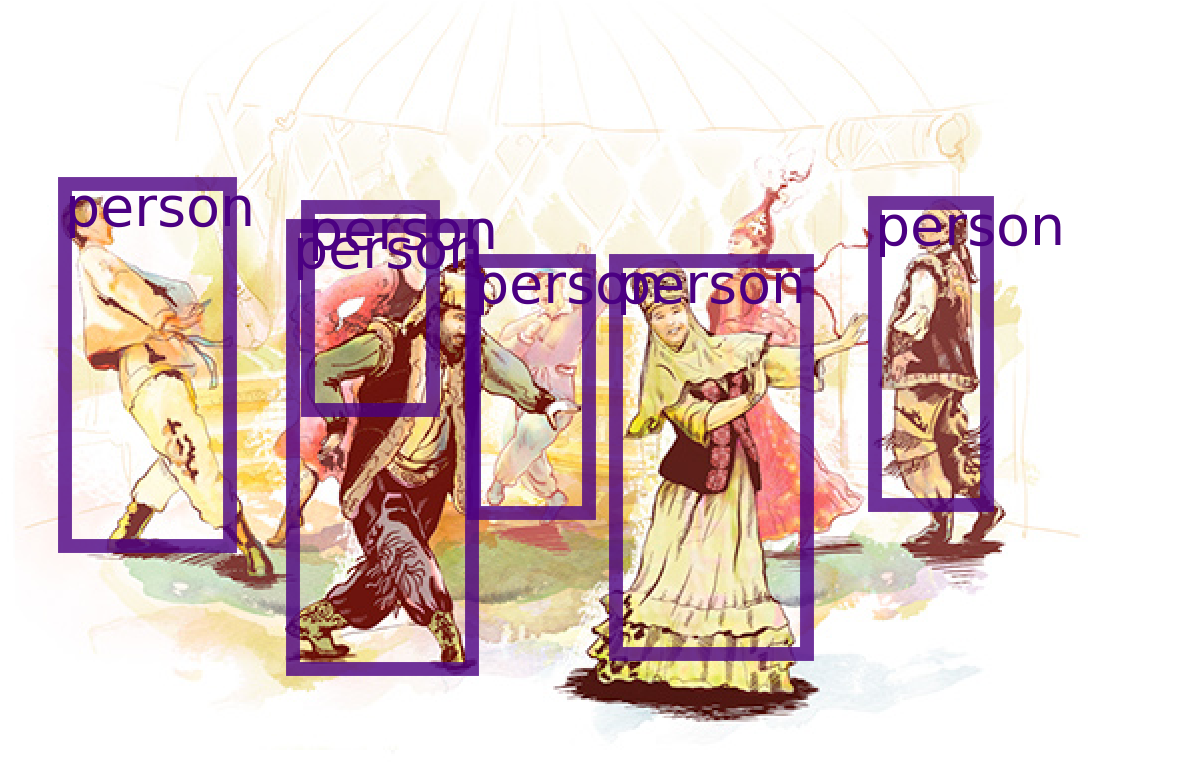}\\
		(a) DAF \cite{DAfasterrcnn} & (b) DAF+SCE \cite{sce} & (c) DAF+CP \cite{cp} & (d) DAF+GCE \cite{gce} & (e) DAF+NLTE (Ours) \\
	\end{tabular}
	\vspace{-6pt}
	\caption{Qualitative results with noisy rate $20\%$ on Clipart1k (top row) and Watercolor2k (bottom row).}\label{fig:results}
	\vspace{-6pt}
\end{figure*}

\begin{figure}[t]
  \centering
    \includegraphics[width=0.99\linewidth]{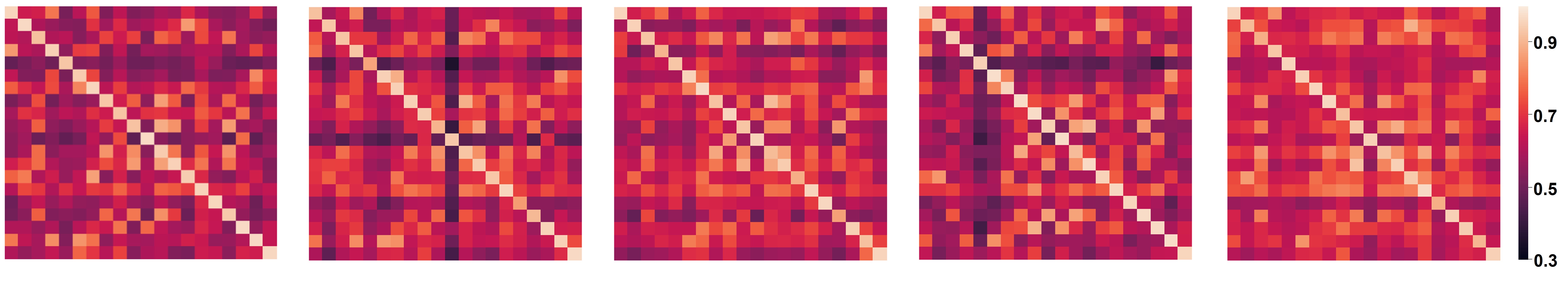}
   \caption{Global relation matrices of Pascal VOC \& Noisy Pascal VOC $\to$ Clipart1k. From left to right refers to noisy rate $0\%$, $20\%$, $40\%$, $60\%$, $80\%$, respectively.}
   \label{fig:relationmatirx}
\end{figure}

\noindent\textbf{Pascal VOC \& Noisy Pascal VOC $\to$ Watercolor2k}.
Table \ref{tab:p2w} shows the experimental results of Pascal VOC and Noisy Pascal VOC $\to$ Watercolor2k. 
In the clean setting, adding all compared methods \cite{sce, cp, gce} perform worse than the baseline DAF \cite{DAfasterrcnn}, indicating that they could suffer from underfitting and hurt the detection performance. 
While the proposed NLTE improves the mAP by $1.3\%$, which is attributed to its capacity of promoting the generalization ability through fully utilizing the noisy samples instead of correcting them straightforwardly.
As the noise rate increases from $20\%$ to $80\%$, adopting NLTE consistently improves the mAP by $2.8\%$, $6.8\%$, $8.4\%$, and $5.5\%$, respectively, while the compared methods show unstable improvements. 
The results evidently suggest that NLTE efficiently boosts the robustness of domain adaptive object detectors.

\subsection{Real-world Noise}
\noindent\textbf{Cityscapes $\to$ Foggy Cityscapes}.
We consider Cityscapes as a real-world noisy annotated source dataset for DAOD and directly implement DAF+NLTE in the Cityscapes $\to$ Foggy Cityscapes benchmark to validate its effectiveness. As listed in Table \ref{tab:cs}, DAF+NLTE shows promising improvements over other state-of-the-arts that were tailored for the DAOD task with the same (or less) training epochs and under the same settings. Specially, with larger training and testing scales as EPM \cite{everypixelmatters} and SSAL \cite{STAL}, \textit{i.e.}, setting the short side of the images to 800 pixels, DAF+NLTE achieves an mAP of $45.4\%$, outperforming SSAL by $5.8\%$. The results indicate that existing DAOD methods may suffer from biased source data and
addressing the noisy annotations is arguably important for achieving effective adaptation.

\begin{figure}[t]
	\centering
	\footnotesize
  \begin{tabular}{p{0.09\textwidth}<{\centering}p{0.08\textwidth}<{\centering}p{0.08\textwidth}<{\centering}p{0.08\textwidth}<{\centering}}
  \multicolumn{4}{c}{\includegraphics[width=6cm]{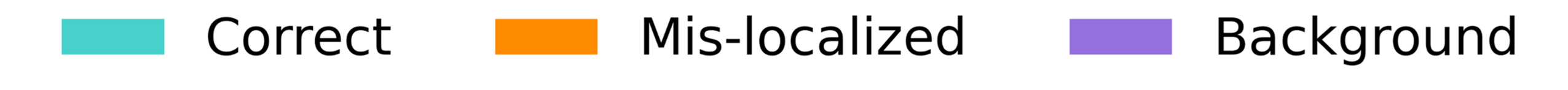}}\\
  \includegraphics[width=1.4cm]{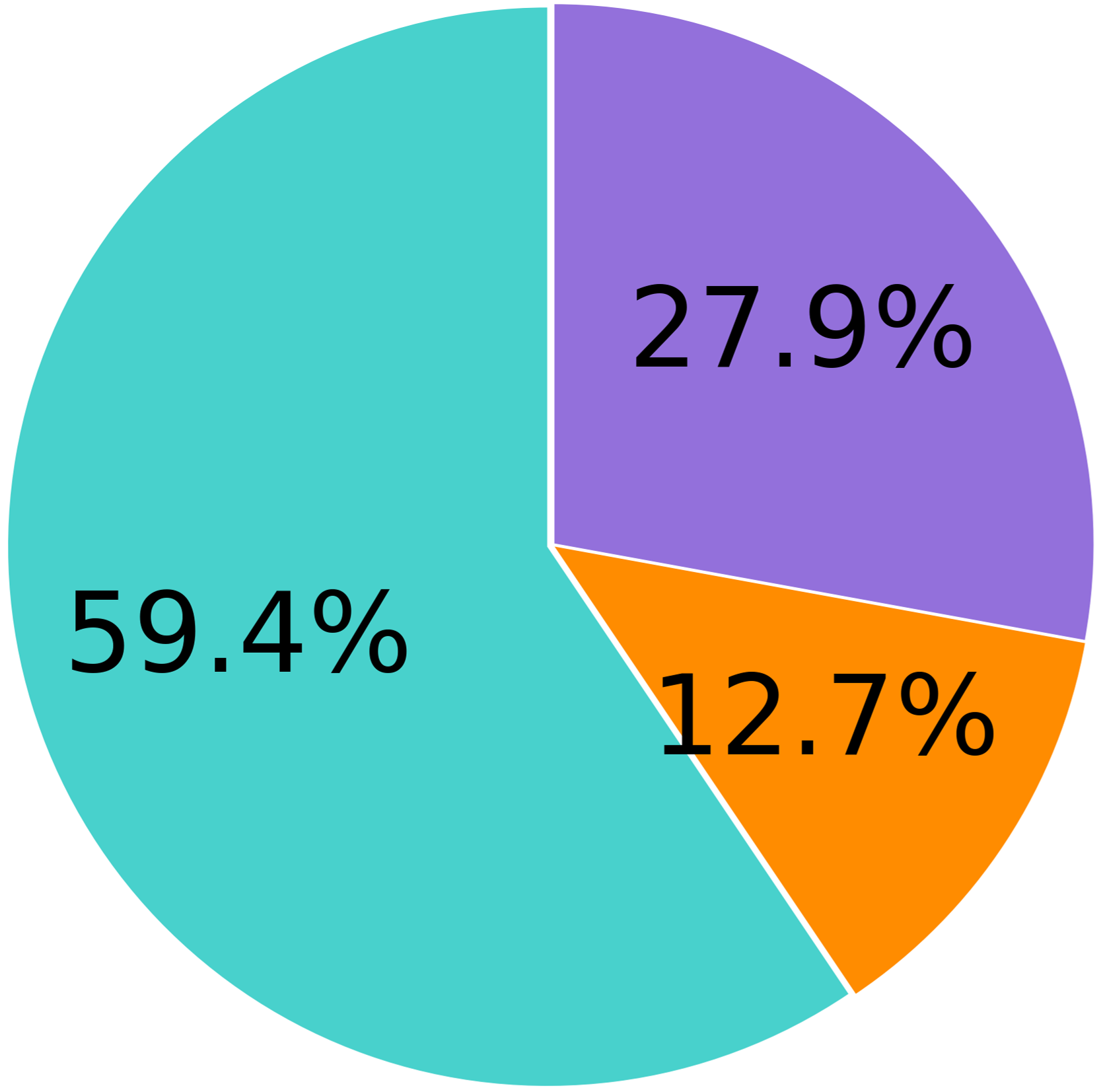}&%
		\includegraphics[width=1.4cm]{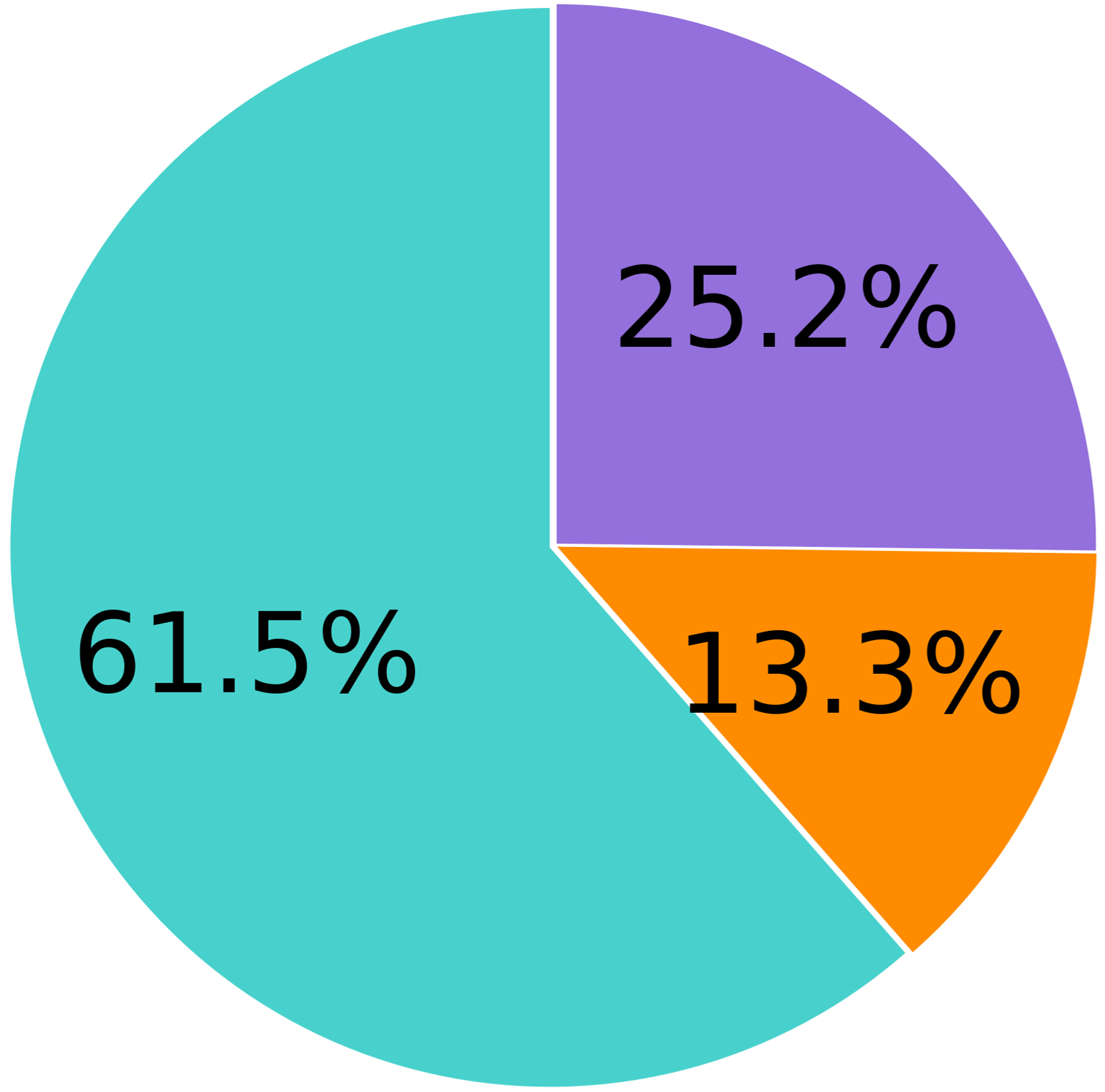}&%
		\includegraphics[width=1.4cm]{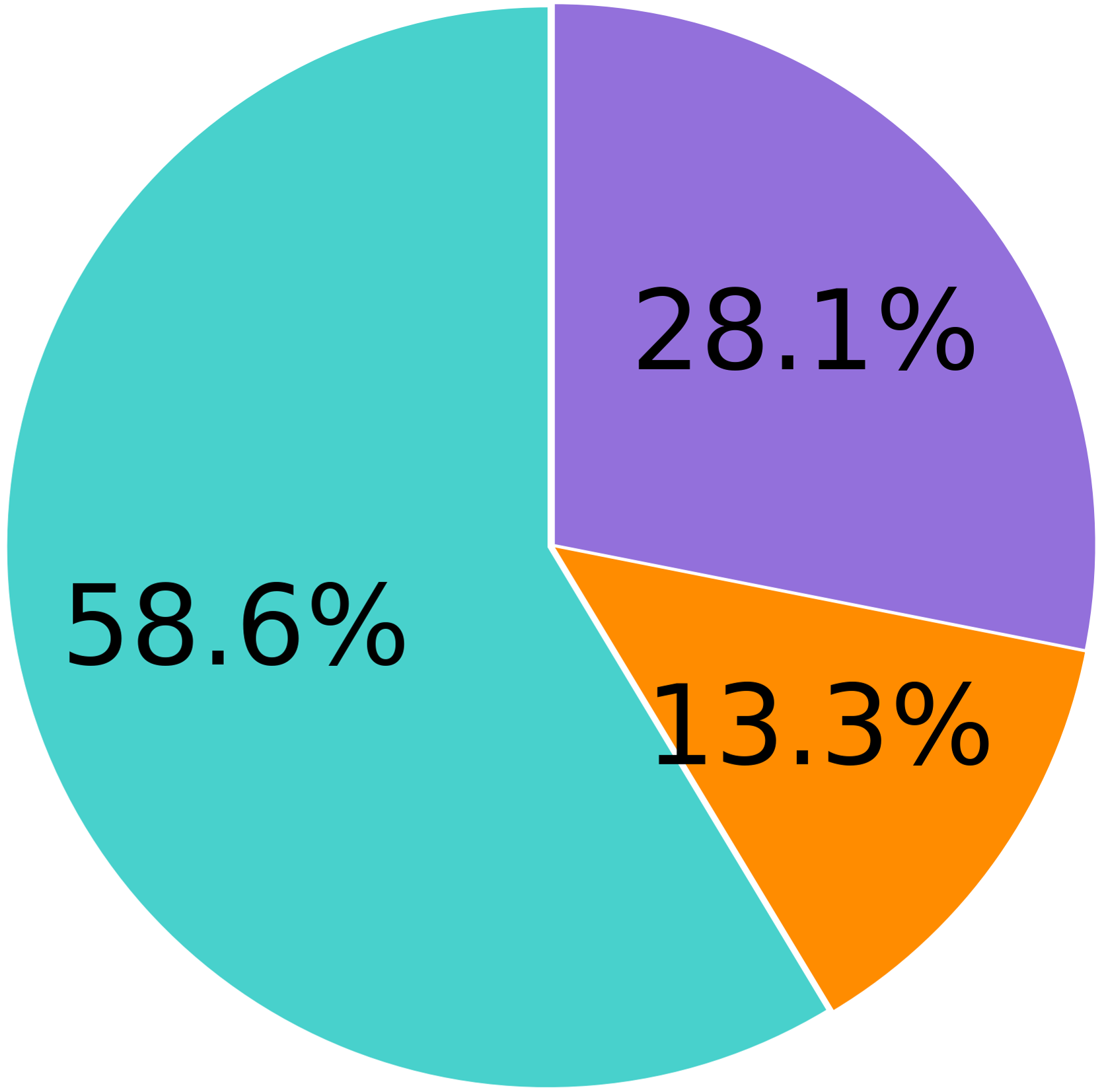}&
		\includegraphics[width=1.4cm]{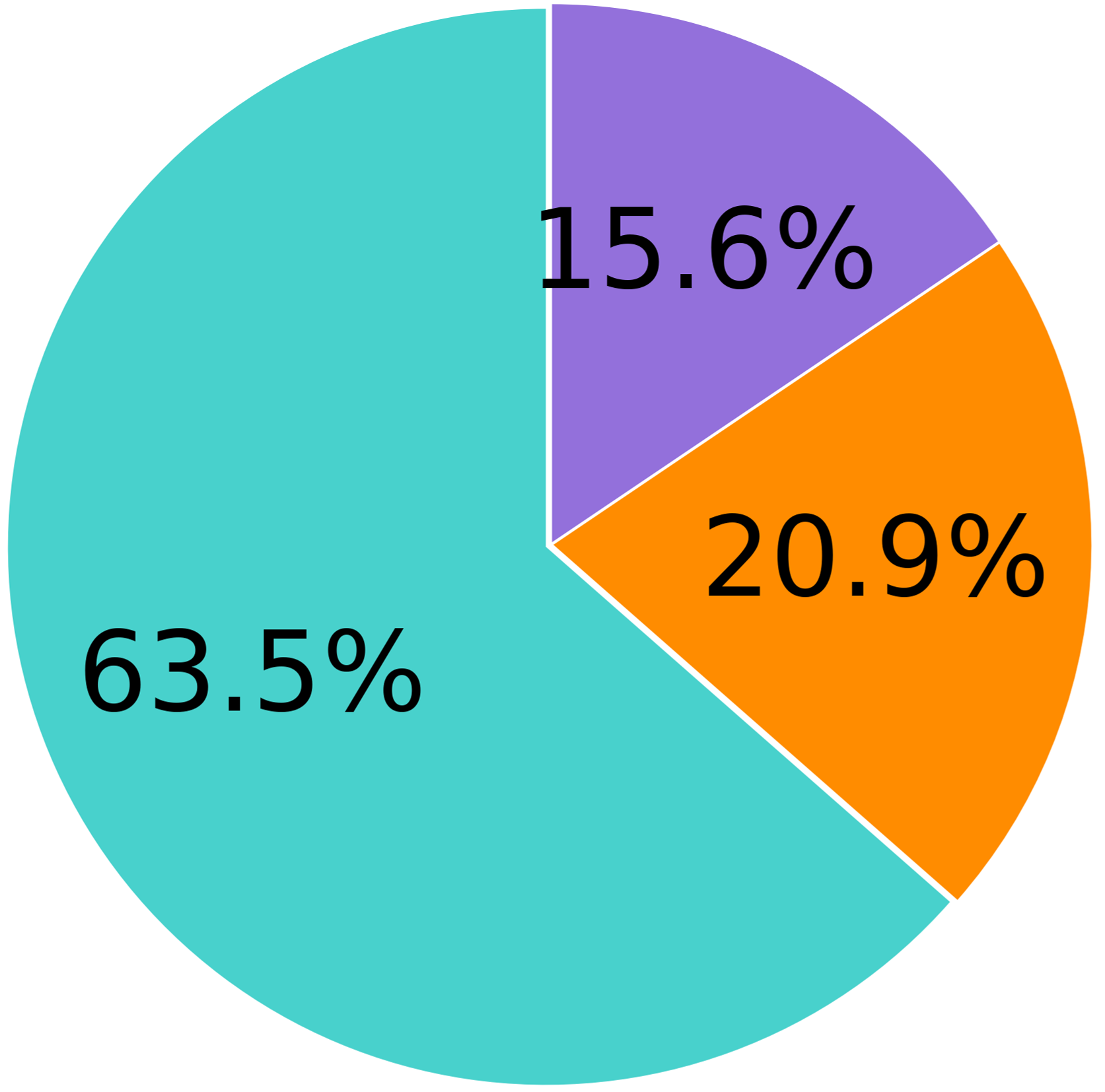}\\
		(a) DAF (40\%) &  (b) +NLTE (40\%) & (c) DAF (60\%) & (d) +NLTE (60\%)\\
	\end{tabular}
	  \caption{Error analysis of highly confident detections on Noisy Pascal VOC $\to$ Watercolor2k. The numbers in brackets refer to noisy rates.}
	\label{fig:pie}
	\vspace{-16pt}
\end{figure}
\vspace{-4pt}
\section{Further Analysis}
\subsection{Ablation Studies} 
Table \ref{tab:ablate} presents the ablation study of the proposed modules in NLTE under small ($20\%$) and large noisy rates ($80\%$) in the Noisy Pascal VOC $\to$ Clipart1k setting. We observe that with PIM, the mAP is improved by $1.1\%$ under $20\%$ noisy rate and $0.3\%$ under $80\%$ noisy rate, indicating that addressing miss-annotation samples in source data would benefit DAOD. With MGRM and EAGR, the mAP is further improved, which demonstrates the effectiveness of utlizing class-corrupted samples in domain alignment. Finally, we demonstrate that strengthening the domain adaptive detector with all components in NLTE boosts the mAP by $2.3\%$ and $4.0\%$ in $20\%$ and $80\%$ noisy rates, respectively, which verifies the effectiveness of NLTE.

\subsection{Qualitative Results}
Fig. \ref{fig:results} illustrates the example of detection results with noisy rate $20\%$. 
From the figure, NLTE can address the semantic confusion problem via PIM and MGRM and correctly classify obscured objects to avoid false positive detections such as \textit{bottle} and \textit{diningtable} even with large domain shift (top row).   
Meanwhile, NLTE can also generate accurate bounding boxes for occluded objects such as \textit{person} and leverage noisy annotated samples to learn domain-invariant representations via EAGR (bottom row). 
\vspace{-2pt}
\subsection{Visualization of the Relation Matrices}
Fig. \ref{fig:relationmatirx} shows the global relation matrices of Pascal VOC \& Noisy Pascal VOC $\to$ Clipart1k. The diagonal entries denote the similarity of the prototypes between the same category and others refer to different categories. It is shown that despite the percentage of noisy annotations increases, global relation matrices can still reflect the class-wise transition probability. With MGRM, the transition probabilities of noisy samples are regularized by global relation matrices, and the latent domain-related knowledge and semantic information are conductive to the domain alignment.
\vspace{-2pt}
\begin{table}[t]
  \centering
  \caption{Ablation studies of the proposed modules in NLTE.}
  \vspace{-8pt}
    \scalebox{0.7}{\begin{tabular}{c|ccc|cc}
    \toprule
    Method & PIM   & MGRM  & EAGR  & 20\%  & 80\% \\
    \midrule
    \midrule
    Baseline \cite{DAfasterrcnn} &       &       &       & 32.8  & 28.5 \\
    +PIM & \checkmark     &       &       & 33.9  & 28.8 \\
    +PIM+MGRM & \checkmark     &  \checkmark    &       & 34.6  & 29.3 \\
    +PIM+EAGR & \checkmark     &       & \checkmark     & 34.5  & 29.5 \\
    +PIM+MGRM+EAGR & \checkmark     & \checkmark     & \checkmark     & \textbf{35.1}  & \textbf{32.5} \\
    \bottomrule
    \bottomrule
    \end{tabular}}%
  \label{tab:ablate}%
  \vspace{-8pt}
\end{table}%

\subsection{Error Analysis of Highly Confident Detections}
To further explore the effect of NLTE, we follow \cite{hoiem2012diagnosing, DAfasterrcnn, mtor, zheng2020cross} to categorize most confident detections into three types: 1) \textbf{Correct} (IoU with GT $\geq$ 0.5), 2) \textbf{Mis-localized}
(0.3 $\leq$ IoU with GT $<$ 0.5), and 3) \textbf{Background} (IoU with GT $<$ 0.3). For each category, we select top-$k$ predictions for analysis, where $k$ is the number of ground truths within the category. Results of mean percentages are shown in Fig. \ref{fig:pie}. On both $40\%$ and $60\%$ noisy rates, NLTE improves the percentage of correct detections and reduces the percentage of false positives. The analysis demonstrates that adopting NLTE could enhance the ability of the detector in distinguishing different classes under noisy scenarios.

\vspace{-6pt}
\section{Conclusion}
In this paper, we address the challenging yet undeveloped issue of domain adaptive object detection under noisy annotations. We propose NLTE, which is a robust adaptive detection framework that simultaneously recaptures miss-annotated samples and explores the transferability of class-corrupted samples. It also harmonizes the gradients between samples for learning domain-invariant representations. Compared with intuitively combining the domain adaptive detector and denoising methods, NLTE shows significant superiority under different noisy rates. Besides, our method outperforms other DAOD methods remarkably in the real-world noise scenario, which implies that addressing the noisy annotations is a suitable and effective alternative to promote the performance of domain adaptive detectors.

{\small
\bibliographystyle{ieee_fullname}
\bibliography{egbib}
}

\end{document}